\documentclass[]{article}
\usepackage{epstopdf}
\usepackage{epsfig}
\begin{document}
\title{A survey of modern optical character recognition techniques}
\date{March 2004}
\author{Eugene Borovikov\\
American Management Systems, Inc.\\
4831 Walden Lane, Lanham, MD 20706\\
+1 (301) 306-2800
}
\maketitle
\pagestyle{myheadings} 
\markright{A survey of modern optical character recognition techniques, AMS 2004}
\begin{abstract}
\noindent This report explores the latest advances in the field of digital document recognition. With the focus on printed document imagery, we discuss the major developments in optical character recognition (OCR) and document image enhancement/restoration in application to Latin and non-Latin scripts. In addition, we review and discuss the available technologies for hand-written document recognition. In this report, we also provide some company-accumulated benchmark results on available OCR engines.
\end{abstract}

\pagebreak
\tableofcontents
\pagebreak
\section{Introduction}

\begin{figure*}
\epsfig{file=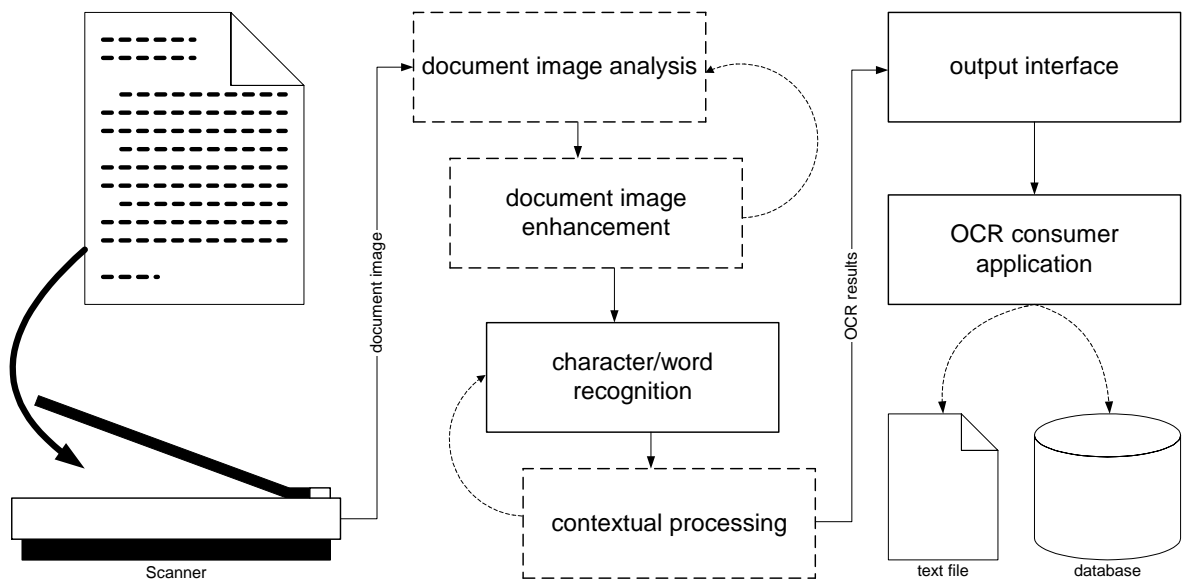,width=\textwidth}
\caption{A typical OCR system}
\label{figOCRStruct}
\end{figure*}

Optical character recognition (OCR) is the process of converting scanned images of machine printed or handwritten text (numerals, letters, and symbols), into machine readable character streams, plain (e.g. text files) or formatted (e.g. HTML files). As shown in Figure~\ref{figOCRStruct}, the data path in a typical OCR system consists of three major stages:
\begin{itemize}
\item document digitization
\item character/word recognition 
\item output distribution
\end{itemize}

In the first stage, the scanner optically captures text in documents and produces document images. Recent advances in scanner technology have made high resolution document scanning widely available. Unlike early black-and-white template matching methods, modern \emph{feature based} optical recognition methods require image spatial resolutions of at least 200 dots per inch (dpi) and can benefit from gray-scale text imagery when it is available. Lower resolutions and simple bi-tonal thresholding tend to break thin lines or fill gaps, thus distorting or invalidating character features needed in the recognition stage.

The second (and the most interesting) stage is responsible for character and/or word recognition in document images. The process involves four operations:
\begin{itemize}
\item optional \emph{image analysis}: image quality assessment, text line detection, word and character extraction, etc.
\item optional \emph{image enhancement}: removing speckle and other image noise, filling holes and breaks, etc.
\item \emph{character/word recognition} usually based on their shapes and other features
\item optional \emph{contextual processing} to limit the feature search space
\end{itemize}

In the third (final) stage, the output interface communicates the OCR results to the outside world. For example, many commercial systems allow recognition results to be placed directly into spread sheets, databases, and word processors. Other commercial systems use recognition results directly in further automated processing and when the processing is complete, the recognition results are discarded. In any case, the output interface is vital to the OCR systems because it communicates results to the user and program domains outside of the OCR system.

\subsection{OCR is a successful branch of Pattern Recognition}
Historically, optical character recognition was considered a technique for solving a particular kind of pattern recognition problems. Indeed, to \emph{recognize} a character from a given image, one would match (via some known metric) this character's \emph{feature pattern} against some very limited \emph{reference set} of known feature patterns in the given alphabet. This clearly is a classical case of a pattern recognition problem.

Character classification has always been a favorite testing ground for new ideas in pattern recognition, but since many such experiments are conducted on isolated characters, the results are not always immediately reflected in OCR applications. Optical character recognition has been steadily evolving during its history, 
giving rise to an exciting set of research topics and producing many powerful practical applications. Now it is considered one of the best applications of machine vision and one of the most successful research branches in pattern recognition theory.

\subsection{Two categories of OCR systems}
Hundreds of OCR systems have been developed since their introduction in the 1950's, and many are commercially available today. Commercial OCR systems can largely be grouped into two categories: \emph{task-specific} readers and \emph{general purpose} page readers.

A \emph{task-specific} reader handles only specific document types. Some of the most common task-specific readers process standard forms, bank checks, credit card slips, etc. These readers usually utilize custom-made image lift hardware that captures only a few predefined document regions. For example, a bank check reader may just scan the courtesy amount field and a postal OCR system may just scan the address block on a mail piece. Such systems emphasize high throughput rates and low error rates. Applications such as letter mail reading have common throughput rates of 12 letters per second with error rates less than 2\%. The character recognizers in many task-specific readers are able to recognize both handwritten and machine-printed text.

\emph{General purpose} page readers are designed to handle a broader range of documents such as business letters, technical writings and newspapers. A typical general purpose reader works by capturing an image of a document page, separating the page into text vs. non-text regions, applying OCR to the text regions and storing non-text regions separately from the output text. Most of the general purpose page readers can read machine written text but only a few can read hand-printed alphanumerics. High-end page readers have advanced recognition capabilities and high data throughput. Low-end page readers usually are compatible with generic flat-bed scanners that are mostly used in an office environment with desk top computers, which are less demanding in terms of system accuracy or throughput. Some commercial OCR software is adaptive and allows users to fine-tune the optical recognition engine to customer's data for improved recognition accuracy.

\subsection{Major trends in modern OCR}
Being a well developed technological field, optical character recognition is widely used today for projects of various scales, from occasional document scanning to creation of massive digital document archives. Yet it remains an area of active scientific research and creative engineering. We can identify the following major research trends in the modern OCR field:
\begin{description}
\item [Adaptive OCR] aims at robust handling of a wider range of printed document imagery by addressing
	\begin{itemize}
	\item multi-script and multi-language recognition~\cite{ChaudhuriICDAR1997OCRSysRd2IndLng,Kanungo1998BibMLOCREvl,Kanungo1999BibMLOCREvl,LuAIPR1999LngIndpOCR,Steinherz1999CrsvScrWrdRec}
	\item omni-font texts~\cite{BazziPAMI1999OmniFontEngAr,JungICDAR1999MltFntClsTypAlg,LeBourgeoisICDAR1997MultiFontOCR,LigatureOCROmnifont,NagySPIE2000OCRIltrGdFrtr}
	\item automatic document segmentation~\cite{CesariniICDAR1999StrDocSegmXYTree,ChowdhuryICDAR2003AutoSegmMathZone,KhedekarICDAR2003TxtImgSep,KornaiICPR1996StcZnFnd,MaCDRR2003BtStrpStrPgSgm,Mukherjee1999DocPgSgmMltScl,NielsonICDAR2003CnsBsdTblFrmRec}
	\item mathematical notation recognition~\cite{ChowdhuryICDAR2003AutoSegmMathZone,GarainICDAR2003MachUndrHnwrMathExpr,JinICDAR2003MathFrmExt,MitraICDAR2003UndrStrcPrntMath,TapiaICDAR2003RecOnlnHwMathEChalk}
	\end{itemize}
\item [Handwriting recognition] \cite{BunkeICDAR2003RecCrvRmnHwr,Steinherz1999CrsvScrWrdRec,SuenICDAR2003AnlRecAsnScr} is a maturing OCR technology that has to be extremely robust and adaptive. In general, it remains an actively researched open problem that has been solved to a certain extent for some special applications, such as
	\begin{itemize}
	\item recognition of hand-printed text in forms~\cite{CharacTell:FormStorm,PegasusImaging:SmartscanXpress,ReadSoft:Forms}
	\item handwriting recognition in personal checks~\cite{KornaiIWFHR1996ChkRcg}
	\item postal envelope and parcel address readers~\cite{KornaiICASSP1997PostalOCR}
	\item OCR in portable and handheld devices~\cite{IBM:HwRecHPC,ClayIrving:HwRecNewton,Microsoft:WinXPTabletPC}
	\end{itemize}
\item [Document image enhancement] \cite{Cannon1999quarc,CherietICDAR2003ShkFltChrImgEnh,LoceDougherty1997,Summers2003ImageRefiner} involves (automatically) choosing and applying appropriate image filters to the source document image to help the given OCR engine better recognize characters and words.
\item [Intelligent post-processing] \cite{Beitzel2002RtrvOCRTxt,CarbonnelICDAR2003LxcPstPrcOptHwr,KleinIROCR2002VotSys4AutOCRCrc,LiACM2002CntxPstPrc,PitrelliICDAR2003CnfScrPstPrcHwr,ZhengICDAR2003TxtIdNsyDocImgMRF} is of great importance for improving the OCR recognition accuracy and for creating robust information retrieval (IR) systems that utilize smart indexing and approximate string matching techniques for storage and retrieval of noisy OCR output texts.
\item [OCR in multi-media] \cite{ClarkBMVC2000FdgTxtRgnLocMsr,LiIP2000AutoTxtDetTrkDigVid,Lienhart2002VideoOCR,Lienhart1998TxtRcgVdNdx,LienhartCSVT2002SgmTxtImgVdWeb,MirmehdiSSPRIP2001ExtLoResTxtAcvCam4OCR,QuickPen:Quicktionary,WuDL1997FdgTxtImg}
is an interesting development that adapts techniques of optical character recognition in the media other than printed documents, e.g. photo, video, and the internet.
\end{description}

\subsection{Main focus and motivation of this report}
The major focus of this survey is on the latest developments in optical character recognition applied to printed document imagery. Cursive and handwriting recognition are also discussed but to a lesser extent. In the subsequent sections, we shall elaborate on the on-going research as well as on commercial and public domain OCR solutions in the context of the topics mentioned above.

More specifically, Section~\ref{secOCR} presents the OCR field with its major techniques, reviews currently available OCR systems, then elaborates on script and language issues, accuracy and performance. Section~\ref{secOCRpreproc} discusses document image defects, importance of document image pre-processing and automatic page segmentation. In Section~\ref{secOCRpostproc}, we address the information retrieval (IR) issues that arise when working with noisy OCR output. We conclude our report with a summary of the discussed OCR issues and with a word on some promising directions that the future may hold for the optical character recognition field.

\section{Optical character recognition}\label{secOCR}

Although OCR stands for ``optical character recognition", it is no mystery that some OCR programs recognize text not by single characters, but rather by character blocks, words, or even groups of words. Therefore, to make the discussion more general it makes sense to talk about \emph{recognition units} or \emph{items}.

\subsection{Essential OCR techniques} 
There exist a wide variety of approaches to optical character recognition. Some of them require pre-segmentation into recognition items, and some of them do not, but we can safely expect that any OCR algorithm would posses these two essential components:
\begin{itemize}
\item a \emph{feature extractor}, and
\item a \emph{classifier}
\end{itemize}
Given an item image, the feature extractor derives the features (descriptors) that the item possesses. The derived features are then used as input to the classifier that determines the (most likely) known item corresponding to the observed features. The classifier is also expected to give a \emph{confidence level} number that tells how certain the classifier is about the recognized item. Let us briefly describe some of the classical optical character recognition methods.
\begin{description}
\item [Template matching] is one of the most common and basic classification methods~\cite[section 8.2]{Parker1997Alg4IPCV}. It is also known as \emph{matrix matching}. A set of all character image prototypes, \emph{templates}, is collected and known a-priori. The feature extractor uses individual image pixels as features. Character classification is performed by comparing an input character image against the template array. Each comparison results in a \emph{similarity measure} (given by a known ``distance" function) between the input character and the template. The similarity increases when a pixel in the observed character is identical to the same pixel in the template image (match). When the corresponding pixels differ (mismatch), the measure of similarity may decrease. As a result, the character's identity is assigned to be the one with the most similar template.
\item [Structural classification] methods employ \emph{structural features} and \emph{decision rules} to classify characters~\cite[section 8.3]{Parker1997Alg4IPCV}. Structural features may be defined in terms of character strokes, character holes, or other character attributes such as corners and concavities. For example, the letter ``P" may be described as a ``vertical stroke with a curved stroke attached on the upper right side making up a hole". To classify the character, the input character structural features are extracted and a rule-based system is applied to compute the character's class.
\end{description}

Note that the above techniques are described in their canonical form, but they have many methodological variations and hybrid approaches. For instance, template matching does not have to be pixel-based, e.g. one could match wavelet transforms of images and templates. Structural classification does not always rely on decision rules, e.g. one could employ a nearest neighbor classifier with an appropriate metric in the feature space.

Many modern OCR systems are based on mathematical formalisms that minimize some measure of misclassification. Such character recognizers may use pixel-based features or structural features. Let us mention some of them:
\begin{description}
\item [Discriminant function classifiers] 
use hypersurfaces in multi-dimensional feature spaces to separate the feature description of characters from different semantic classes. Such classifiers aim at reducing the \emph{mean-squared classification error}.
\item [Bayesian classifiers] 
seek to minimize some \emph{loss function} associated with character misclassification through the use of probability theory.
\item [Artificial Neural Nets] 
(ANN) originating from the theories of human and animal perception, employ \emph{error back-propagation} techniques to learn some non-trivial character classification maps and use mathematical optimization techniques to minimize possible classification errors.
\end{description}

\subsection{Currently available OCR systems}\label{secAvlOCRSys}
Since the introduction of the first OCR system, there have been hundreds if not thousands of different OCR systems both commercial and public domain. Below, we mention and briefly characterize some currently available OCR engines.
\subsubsection{Commercial OCR solutions} 
Many commercial OCR systems started as university projects, shareware or freeware programs, and developed into sophisticated high quality OCR engines, software development kits (SDK), and ready-to-use software products meeting the high expectations of today's OCR market.
\begin{description}
\item [Capture Development System 12] by ScanSoft ({\tt http://www.scansoft.com/}) allows software developers to reduce the cost of development and time-to-market by providing accurate OCR engines, advanced image enhancement tools, document processing capabilities and support for a wide range of input/output filters, including PDF, XML and Open eBook standards. The standard version handles European scripts, and has an extension that handles Chinese, Japanese, and Korean respectively.
\item [FineReader 7.0] by ABBYY ({\tt http://www.abbyy.com/}) is an OCR application designed specifically for network environments. It combines a high level of accuracy and format retention with robust networking capabilities, providing a good OCR solution for corporations that need to convert and reproduce a variety of paper documents and PDF files for editing and archiving purposes. FineReader reads 177 languages, including English, German, French, Greek, Spanish, Italian, Portuguese, Dutch, Swedish, Finnish, Russian, Ukrainian, Bulgarian, Czech, Hungarian, Polish, Slovak, Malay, Indonesian, and others. Built-in spell-check is available for 34 languages.
\item [Automatic Reader 7.0] by Sakhr ({\tt http://www.sakhr.com/}) is one of the standard OCR readers for Arabic script. Sakhr OCR combines two main technologies: Omni Technology, which depends on highly advanced research in artificial intelligence, and Training Technology, which increases the accuracy of character recognition. It can perform automatic document image segmentation and
identify multiple scripts through Xerox Text Bridge technology.
\item [NovoD DX] by Novodynamics ({\tt http://www.novodynamics.com/}) is an integrated information retrieval (IR) system that can accurately handle large volumes of documents written in the targeted languages. The system is able to
\begin{itemize}
\item accurately handle degraded documents
\item handle documents written in Asian languages, e.g. Arabic
\item be extended to a variety of languages
\end{itemize}
NovoDynamics initially designed this software to work with certain languages as part of a program with the Department of Defense, and it was enhanced through the business collaboration with In-Q-Tel.
\item [NeuroTalker] by International Neural Machines can be used as a ``fast, lightweight" OCR engine that operates by copying an image to the clipboard and then pasting it as text.
\item [BBN] is a trainable system that claims to be completely script-independent, other than the fact that training and application data should obviously match.
\item [Prime OCR] is a voting system that one can acquire with up to 6 other commercial engines inside it.
\item [Cuneiform] is claimed to be a fast OCR engine for Latin scripts.
\item [Vividata OCR] software by Vividata utilizes Caere (ScanSoft) OCR technologies in several LINUX and UNIX platforms.
\end{description}

\subsubsection{Public domain OCR resources} 
Most publicly available (free) OCR software has been developed primarily for UNIX and LINUX based systems~\cite{CfAR:FreeOCR}. The public domain OCR programs are written in most popular programming languages, such as C or C++, and are distributed along with their source code, thus they are \emph{open source} systems. Besides being free, the open source OCR systems are extremely flexible and software developer friendly. One obvious drawback of open source systems is the need to compile and make them executable on the user's local platform, which requires an appropriate compiler and often the expertise of a software developer. Below we give a list of a few such systems:
\begin{description}
\item [NIST Form-Based Handprint Recognition System:]~\cite{Garris1994NISTFrmHndPrtRecSys,Garris1997NISTFrmHndPrtRecSys2} a standard reference form-based handprint recognition system developed at NIST. The recognition system processes the Handwriting Sample Forms distributed with NIST Special Databases 1 and 3. The system reads handprinted fields containing digits, lower case letters, upper case letters, and reads a text paragraph containing e.g. the Preamble to the U.S. Constitution. Source code is provided for form registration, form removal, field isolation, field segmentation, character normalization, feature extraction, character classification, and dictionary based post-processing. The system is known to have been successfully compiled and tested on a Digital Equipment Corporation (DEC) Alpha, Hewlett Packard (HP) Model 712/80, IBM RS6000, Silicon Graphics Incorporated (SGI) Indigo 2, SGI Onyx, SGI Challenge, Sun Microsystems (Sun) IPC, Sun SPARCstation 2, Sun 4/470, and a Sun SPARC-station 10.
\item [Illuminator:]~\cite{RAF:IllumDAFS} an open source software toolset for developing OCR and Image Understanding applications developed by RAF Technology in 1999. Illuminator has two major parts: a library for representing, storing and retrieving OCR information, referred to as DAFSLib, and an X-Windows ``DAFS" file viewer, called illum.
\item [CalPoly OCR:]~\cite{CalPolyOCR} software originally written under the supervision of Prof. Clint Staley for a Computer Vision class project at California Polytechnic State University. The system is known to be font-specific, working well on fonts from the high-end HP laser printer on which it was trained, but is not guaranteed to work on others. It was developed at Cal Poly using university resources, and thus may be used only for academic purposes, not for commercial gain.
\item [Xocr:]~\cite{Bauer:Xocr} a shareware OCR software written by Martin Bauer ({\tt Martin\_Bauer@S2.maus.de}) for X11 on a LINUX system. With some effort it can be ported to SunOS.
\end{description}

\subsection{Why optical character recognition is difficult} 
Character misclassifications stem from two main sources: poor quality recognition unit (item) images and inadequate discriminatory ability of the classifier. There are many factors that contribute to noisy, hard to recognize item imagery:
\begin{itemize}
\item poor original document quality
\item noisy, low resolution, multi-generation image scanning
\item incorrect or insufficient image pre-processing
\item poor segmentation into recognition items
\end{itemize}

On the other hand, the character recognition method itself may lack a proper response on the given character (item) set, thus resulting in classification errors. This type of errors can be difficult to treat due to a limited training set or limited learning abilities of the classifier.

Typical recognition rates for machine-printed characters can reach over 99\% but handwritten character recognition rates are invariably lower because every person writes differently. This random nature often manifests itself in a greater character variance in the feature space leading to greater misclassification rates.

A common example of a ``difficult" character is the letter ``O" easily confused with the numeral ``0". Another good example could be the letter ``l" confused with the numeral ``1" or mistaken for a noisy image of the letter ``I".

Rice et al.~\cite{NagySPIE2000OCRIltrGdFrtr,Rice1999OCRGdFrtr} discuss the character recognition abilities of humans vs computers and present illustrated examples of recognition errors. The top level of their taxonomy of error causes consists of
\begin{itemize}
\item \emph{imaging defects} due to heavy/light print, stray marks, curved baselines, etc.
\item \emph{similar symbols} as mentioned above
\item \emph{punctuation} due to commas and periods, quotation marks, special symbols, etc.
\item \emph{typography} due to italics and spacing, underlining, shaded backgrounds, unusual typefaces, very large/small print, etc.
\end{itemize}
Their analysis provides insight into the strengths and weaknesses of current systems, and a possible road map to future progress. They conclude that the current OCR devices cannot read even on the level of a seven-year old child. The authors consider four potential sources of improvement:
\begin{itemize}
\item \emph{better image processing} based on more faithful modeling of the printing, copying and scanning processes
\item \emph{adaptive character classification} by fine-tuning the classifier to the current document's typeface
\item \emph{multi-character recognition} by exploiting style consistency in typeset text
\item \emph{increased use of context} that depends on the document's linguistic properties and can vary from language to language
\end{itemize}
On the basis of the diversity of errors that they have encountered, they are inclined to believe that further progress in OCR is more likely to be the result of multiple combinations of techniques than on the discovery of any single new overarching principle.


\subsection{Document image types}
Most optical character recognition systems are designed to deal with bi-tonal images of black text on white background. Even in this seemingly simple imaging terrain, there exist a variety of image formats due to variations in image spatial resolutions, color encoding approaches and compression schemes.

Typical facsimile scanned document images are bi-tonal (one bit per pixel) with spatial resolutions of 200 dots per inch (dpi). This resolution may not always be sufficient for high-quality OCR. Modern character recognition engines work better with documents scanned at 300 dpi or higher. Digital libraries trying to reasonably preserve the original document quality, often require gray-scale (8 bits per pixel) or even true-color (24 bits per pixel) scanning at 600 dpi.

High-resolution color images, however, take their toll on a system's storage, and can become prohibitively large to store in large quantities. Smart digital libraries use different \emph{compression} schemes for text and non-text content. A typical compression employed for bi-tonal text images is CCITT Fax Group 3 and Group 4~\cite{CCITT:T.4,CCITT:T.5,CCITT:T.6}, while real-world imagery is best compressed with JPEG technology. Below we characterize some of the most popular image and file formats and their applicability to OCR.
\begin{description}
\item [TIFF] stands for Tagged Image File Format. It is one of the most popular and flexible of the current public domain raster file formats designed for raster data interchange. TIFF was developed jointly by Aldus and Microsoft Corporation. Adobe Systems acquired Aldus and now holds the Copyright for the TIFF specification. Since it was designed by developers of printers, scanners and monitors, it has a very rich space of information elements for colorimetry calibration, and gamut tables.

Theoretically, TIFF can support imagery with multiple bands (up to 64K bands), arbitrary number of bits per pixel, data cubes, and multiple images per file, including thumbnail sub-sampled images. Supported color spaces include gray-scale, pseudo-color (any size), RGB, YCbCr, CMYK, and CIELab. TIFF supports the following compression types: raw uncompressed, PackBits, Lempel-Ziv-Welch (LZW), CCITT Fax 3 and 4, JPEG; and pixel formats: 1-64 bit integer, 32 or 64 bit IEEE floating point.

TIFF's main strength is that it is a highly flexible and platform-independent format that is supported by numerous image processing applications. One of the main limitations of TIFF is the lack of any provisions for storing vector graphics and text annotation, but this is not very threatening to OCR since most recognition engines work with raster image representation.

\item [BMP] or \emph{bitmap}, is a raster image format widely used in the PC world under Windows and OS/2 operating systems. BMP files are stored in a device-independent bitmap (DIB) format that allows the operating system to display the bitmap on any type of display device. DIB specifies pixel colors in a form independent of the method used by a display to represent color.

Windows versions 3.0 and later support run-length encoded (RLE) formats for compressing bitmaps that use 4 or 8 bits per pixel. Compression reduces the disk and memory storage required for a bitmap. Evidently, this compression comes at the expense of the color variety and depth, but this does not present a problem for bi-tonal or gray-scale based OCR systems.

Bitmap files are especially suited for the storage of real-world images; complex images can be rasterized in conjunction with video, scanning, and photographic equipment and stored in a bitmap format. BMP is a straight-forward image format that is perfectly suitable for storing document images, but it is not as elaborate and flexible as TIFF.

\item [PCX] is one of the oldest raster formats available on PC's and was originally established by Z-soft for its PC based Paintbrush software. Because it has been around for such a long time, there are now many versions of PCX. Most software today supports version 5 of the PCX format. Version 3 only supports 256 colors, but it does not allow for a custom palette. This means that when you open a version 3 PCX file, a standard VGA palette is used instead.  PCX retains all image information (similar to BMP) but it uses no compression, and hence can be memory-inefficient.

\item [JPEG] stands for Joint Photographic Experts Group, the original name of the
committee that developed this standard for \emph{lossy image compression} that sacrifices some image quality to considerably shrink the file size. JPEG was designed for compressing either full-color or gray-scale images of natural, real-world scenes. It works well on photographs, naturalistic artwork, and similar material; not so well on lettering, simple cartoons, or line drawings. JPEG compression exploits the fact that human eyes perceive small color changes less accurately than small changes in brightness. Thus, JPEG was intended for compressing images that will be looked at by humans. However, small errors introduced by JPEG compression may be problematic for machine vision systems; and given its less than adequate performance on bi-tonal imagery, JPEG is not a recommended image file format in OCR.

\item [GIF] stands for Graphic Interchange Format. It was popularized by the CompuServe Information Service in the 1980s as an efficient means to transmit images across data networks. In the early 1990s, GIF was adopted by the World Wide Web for its efficiency and widespread familiarity. Today the overwhelming majority of images on the Web are in GIF format. GIF files are limited to 8-bit color palettes supporting no more than 256 colors. The GIF format incorporates a compression scheme to keep file sizes at a minimum. GIF works for images of text and diagrams better than for real-world images. This makes GIF an attractive (and often more straight-forward than TIFF) choice for OCR systems that work with bi-level and gray-scale images. TIFF's CCITT Fax compression, however, produces more compact images.

\item [PNG] stands for Portable Network Graphics. This format was designed to phase out the older and simpler GIF format, and to provide a practical alternative to the much more complex TIFF format.

For the Web, PNG really has three main advantages over GIF: alpha channels (variable transparency), gamma correction (cross-platform control of image brightness), and two-dimensional interlacing (a method of progressive display). As does GIF, PNG provides a \emph{lossless} compression (that does not sacrifice image quality for smaller file size), but does a better job at it. Unlike GIF, PNG supports only a single image per file.

Unlike TIFF, the PNG specification is strict on implementation of supported format features, which translates into a much better (than TIFF) image file portability. However, TIFF's Group 4 fax compression or the JBIG format are often far better than 1-bit gray-scale PNG for black-and-white images of text or drawings.

\item [PDF] stands for Portable Document Format. It was developed by Adobe Corporation to capture exact formatting information from document publishing applications, making it possible to exchange formatted documents and have them appear on the recipient's monitor or printer as they were intended to look in the original application.

Being a compact version of the PostScript format, PDF is a very advanced and elaborate file format that supports both vector and raster graphics, applying various lossy and lossless compression schemes to reduce file size. Although this is a proprietary Adobe format, both the file format specifications and the viewer software are freely distributed. PDF images are moderate in size, typically around 50KB per page at 300 dpi. Typical applications for this format include cost-effective document archiving (where text-searchability is not required) and print/fax on demand. To work with PDF images of text, OCR software needs a sophisticated PDF reader that can be purchased or built by the freely available format specifications.
\end{description}

This rich variety of image formats can be overwhelming to support in a single OCR system, and image format converters often have to be employed. Such converters may be available as
\begin{itemize}
\item stand-alone (sometimes open-source) utility programs (e.g. bmp2tiff)
\item part of graphics libraries (e.g. ImageMagick)
\item part of commercial image processing software (e.g. Adobe Photo Shop)
\end{itemize}

Although concerned primarily with scanned paper documents, OCR technology is currently experiencing expanded applicability to other media such as digital photo, video, and the world wide web. The primary task there remains the same: finding text in images; but the image format does not have to be fixed to black-and-white. Working in this challenging area, researchers pursue the ultimate goal of building an OCR system that can recognize any text in any still or moving picture encoded in any format.

In the following two subsections, we discuss image formats that are traditional for OCR (scanned documents) and not so traditional for it (other media).

\subsubsection{Scanned documents}
Due to such a variety in document image formats, most commercial-off-the-shelf (COTS) OCR solutions work with more than a single image file format and can adapt to various spatial resolutions and pixel depths. Many COTS OCR packages come with a rich set of image processing utilities that analyze and transform input images to the format most suitable for the given OCR engine. Let us mention some popular commercial OCR solutions along with the image formats they support.
\begin{description}
\item [ScanSoft Capture Development System 12] features a wide range of image and application format support, including BMP, GIF, TIF, PDF, HTML, Microsoft Office formats, XML, and Open eBook.
\item [ABBYY Fine Reader 7.0] can work with black-and-white, gray-scale and color images in various formats including BMP, PCX, DCX, JPEG, JPEG 2000, PNG, PDF, and TIFF (including multi-image, with the following compression methods: Unpacked, CCITT Group 3, CCITT Group 4 FAX(2D), CCITT Group4, PackBits, JPEG, ZIP)
\item [Sakhr Automatic Reader 7.0] provides necessary functionality to read and write images in various formats including TIFF, PCX, BMP, WPG, and DCX.
\end{description}
Most consumer-level OCR programs work with bi-level imagery and typically expect a black text on a white background. Some can import gray-scale images and internally convert them to black-and-white, sometimes using adaptive thresholding. Only rare OCR engines can directly work with gray-scale or color images taking advantage of multi-bit pixel information.

\subsubsection{Other media}
Many multimedia sources provide images and video that contain visible text that can potentially be detected, segmented, and recognized automatically. Such capability would be much desirable for creating advanced multimedia information retrieval systems. The text captured from multimedia content would be a valuable source of high-level semantic information for indexing and retrieval.

One of the most interesting pieces of research in this direction is carried out by Lienhart et al.~\cite{Lienhart2002VideoOCR,Lienhart1998TxtRcgVdNdx,LienhartCSVT2002SgmTxtImgVdWeb}. They propose an interesting method for localizing and segmenting text in complex images and videos. Text lines are identified by using a complex-valued multi-layer feed-forward network trained to detect text at a fixed scale and position. The network's output at all scales and positions is integrated into a single text-saliency map, serving as a starting point for candidate text lines. In the case of video, these candidate text lines are refined by exploiting the temporal redundancy of text in video. Localized text lines are then scaled to a fixed height of 100 pixels and segmented into a binary image with black characters on white background. For videos, temporal redundancy is exploited to improve segmentation performance. Input images and videos can be of any size due to a true multi-resolution approach. The presented method appears to be a generic, scale-invariant solution for text localization and text segmentation in images and videos. On a real-world test set of video frames and web pages, 69.5\% of all text boxes were located correctly. The
performance rose to 94.7\% by exploiting the temporal redundancy in videos. 79.6\% of all characters in the test video set could be segmented properly and 88\% of them were recognized correctly.


\subsection{Script and language issues}
We define \emph{script} as the way to write in a particular \emph{language}. It is not uncommon for multiple languages to share same or similar scripts, e.g.
\begin{itemize}
\item most Western-European languages sharing variations of the Latin script
\item some Eastern-European languages sharing variations of the Cyrillic script
\item many Islamic culture languages sharing variations of the Arabic script
\end{itemize}
Therefore, having developed a system for one script, it is possible to easily adapt it to a whole family of languages. Until the early 1990's, the optical character recognition technology was (with rare exceptions) concerned with the Latin script. This resulted in a rich variety of Latin-based OCR systems. With the global expansion of information technology, recognition of scripts other than Latin now becomes a worldwide interest.

\subsubsection{Complex character scripts}
As we mentioned earlier, OCR for Latin-like scripts is fairly well researched and has numerous successful implementations, while most Asian scripts are very different from the Latin script and thus present different challenges to the OCR community.

\begin{figure}[h]
\centering
\epsfig{file=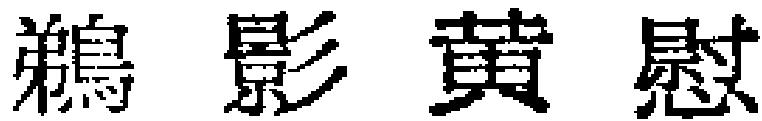}
\caption{Complex Kanji characters}
\label{figCplxKanji}
\end{figure}

\begin{figure}[h]
\centering
\epsfig{file=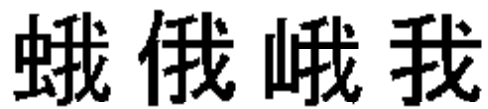}
\caption{Many characters share the same lexicographical element}
\label{figSameLexEl}
\end{figure}

Let us, for instance, consider the Japanese script. In addition to alphanumerics, Japanese text uses Kanji characters (Chinese ideographs) and Kana (Japanese syllables). Therefore, it is conceivably more difficult to recognize Japanese text because of 
\begin{itemize}
\item the size of the character set (exceeding 3300 characters),
\item the complexity of individual characters (Figure~\ref{figCplxKanji}), and
\item the similarities between the Kanji character structures (Figure~\ref{figSameLexEl}).
\end{itemize}

Low data quality is an additional problem in all OCR systems. A typical Japanese OCR system is usually composed of two individual classifiers (pre-classifier and secondary classifier) in a cascade structure. The pre-classifier first performs a fast coarse classification to reduce the character set to a short candidate list (usually contains no more than 100 candidates). The secondary classifier then uses more complex features to determine which candidate in the list has the closest match to the test pattern.

\begin{figure}[h]
\centering
\epsfig{file=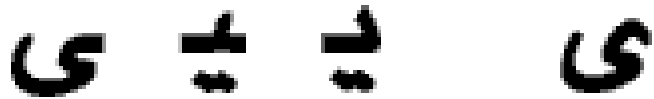}
\caption{Four shapes of the Arabic letter ``yeh" (from right to left): isolated, initial, medial, and final}
\label{figArbY}
\end{figure}

Arabic character recognition has its challenges, too. To begin with, Arabic text is written, typed and printed cursively in blocks of interconnected characters. A word may consist of several character blocks. Arabic characters in addition to their isolated form can take different shapes depending on their position inside the block of characters (initial, medial, or final, as in Figure~\ref{figArbY}).

\begin{figure}[h]
\centering
\epsfig{file=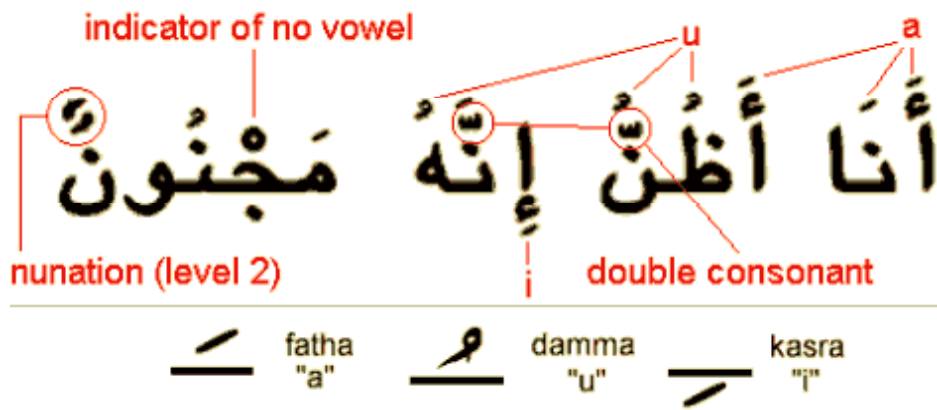}
\caption{Arabic diacritics}
\label{figArbDcrt}
\end{figure}

Arabic characters can also be written stacked one on top of another, which may lead to character blocks having more than one base line. Additionally, Arabic uses many types of external objects such as dots, \emph{Hamza} and \emph{Madda}. Optional diacritics (shown in Figure~\ref{figArbDcrt}) also add to the set of external objects.

Finally, Arabic font suppliers do not follow a common standard. Given the peculiarities of Arabic fonts and the characteristics of the Arabic language, building an omni-font Arabic OCR becomes a difficult undertaking.

\subsubsection{Techniques for handling various scripts}
Steinherz et al.~\cite{Steinherz1999CrsvScrWrdRec} review the field of off-line cursive word recognition. They classify the field into three categories:
\begin{itemize}
\item \emph{segmentation-free methods}: compare a sequence of observations derived from a word image with similar references of words in the lexicon,
\item \emph{segmentation-based methods}: look for the best match between consecutive sequences of primitive segments and letters of a possible word,
\item \emph{perception-oriented approach}: perform a human-like reading technique, in which anchor features found all-over the word are used to boot-strap a few candidates for a final evaluation phase.
\end{itemize}
In their view the field is not mature enough to enable a comparison of the above methods. It appears to them that the problem of word recognition in the case of small and static lexicons is essentially solved, but the same is not the case for limited and large lexicons. They would like to point out that the current systems may be close to the limits of stand-alone word recognition, and efforts should be made to improve post-processing techniques that will take advantage of syntax, context, and other external parameters.

Khorsheed et al.~\cite{KhorsheedPAA2002OflACRRv,Khorsheed1999ArbWrdRecHMM,KhorsheedBMVC1999StrcFtrCrsvArbScrp} review various techniques in Arabic character recognition and focus on a word-based technique for extracting structural features from cursive Arabic script. Their method skeletonizes word images and decomposes them into a number of segments that are used to form a feature vector that in turn is used to train the Hidden Markov Model to perform the recognition. This method depends highly on a predefined lexicon which acts as a look-up dictionary. With the experimental word model trained from a $294$-word lexicon acquired from a variety of script sources, their system is claimed to have achieved recognition rates of up to $97$\%.

Lu et al.~\cite{LuAIPR1999LngIndpOCR} present a language-independent OCR system that  they claim to be capable of recognizing printed text from most of the world's languages. The system uses Hidden Markov Models (HMM) for character recognition, but does not require source data pre-segmentation into characters or words. Training and recognition are performed using an existing continuous speech recognition system. The system uses \emph{unsupervised adaptation} techniques to increase its robustness to image degradation due to e.g. facsimile. To demonstrate the system's language-independence, the authors have experimented with three languages: Arabic, English, and Chinese. The Chinese OCR system can recognize $3870$ characters, including $3755$ simplified Chinese characters. The Arabic and English systems provide high quality unlimited vocabulary omni-font recognition. The authors also demonstrate that the usage of a lexicon can significantly improve the character error rate of their OCR system.

\subsection{Effects of language mixing}
There has been a growing interest in multi-lingual and multi-script OCR technology during recent years. Multi-lingual OCR systems are naturally designed to read more than one script in the same document, and are especially important for document exchange and business transactions in many countries across Europe and Asia, where documents are often printed in multiple languages.

The ability to reliably identify scripts using the least amount of textual data is essential when dealing with document pages that contain multiple languages. There are several challenges to the script identification task, e.g.

\begin{itemize}
\item unexpected script or language is always a possibility, which presents a problem in the choice of an OCR engine down the pipeline
\item no character identification allowed, since script analysis takes place prior to the actual text recognition
\item processing time of script identification should be a small fraction of the total OCR processing time
\end{itemize}

Hochberg et al.~\cite{Hochberg1997PgSgmScrIdVct,HochbergPAMI1997AutoScrIdClstr} 
describe an automated script identification system that uses \emph{cluster-based templates} in the analysis of multilingual document images. A \emph{script identification vector} is calculated for each connected component in a document. The vector expresses the closest distance between the component and templates developed for each of 13 scripts (Arabic, Armenian, Burmese, Chinese, Cyrillic, Devanagari, Ethiopic, Greek, Hebrew, Japanese, Korean, Roman, and Thai.) The authors calculate the first three principal components within the resulting thirteen-dimensional space for each image. By mapping these components to red, green, and blue, they visualize the information contained in the script identification vectors. The visualizations of several multilingual images suggest that the script identification vectors can be used to segment images into script-specific regions as large as several paragraphs or as small as a few characters. The visualized vectors also reveal distinctions within scripts, such as font in Roman documents, and kanji vs. kana in Japanese. Their experiments were run on a limited number of multi-script documents scanned at 200 dpi. Results appear to be best for documents containing highly dissimilar scripts such as Roman and Japanese. Script identification on documents containing similar scripts, such as Roman and Cyrillic, will require further investigation.

Current approaches to script identification often rely on hand-selected features and often require processing a significant part of the document to achieve reliable identification. Ablavsky and Stevens~\cite{AblavskyICDAR2003AutoFtrSelScrId} present a shape-based approach that applies a large pool of image features to a small training sample and uses subset feature selection techniques to automatically select a subset with the most discriminating power. At run time they use a classifier coupled with an evidence accumulation engine to report a script label once a preset likelihood threshold has been reached. Algorithm validation took place on a corpus of English and Russian document images with 1624 lines of text. The sources were newsprint and books scanned at 300 dpi.
\begin{table*}[h]
\centering
\begin{tabular}{|c|c|c|c|}
\hline
 & ambiguous & Cyrillic & Latin \\
\hline
Cyrillic & 0.05 & 0.94 & 0.01 \\
Latin & 0.14 & 0.42 & 0.43 \\
\hline
\end{tabular}
\caption{Confusion matrix for automatic script identification}
\label{tblCyrLatCnfMx}
\end{table*}
Degraded documents were first-generation photocopies of first-generation facsimiles of photocopies (Table~\ref{tblCyrLatCnfMx}). The major source of errors came from imperfect zoning, inaccurate line extraction in multi-column text, and joined characters. Script classification of degraded but properly segmented characters is claimed to be accurate.

Pal et al.~\cite{PalICDAR2003MltScrLnIdIndDoc} give an automatic scheme to identify text lines of different Indian scripts. Their method groups the scripts into a few classes according to script characteristics. It uses various script features (based on e.g. water reservoir principle, contour tracing and character profile) to identify scripts without any expensive OCR-like algorithms. The method was applied to 250 multi-script document images containing about 4000 text lines. The images were scanned from juvenile literature, newspapers, magazines, books, money order forms, computer printouts, translation books, etc. The authors report that their system has an overall accuracy of about 97.52\%.


\subsection{Character types}
With all the modern advances in the OCR field, it is still challenging to develop a recognition system that can maintain a high recognition rate on the documents that use complex character scripts or a variety of fonts and typefaces. We can identify two major approaches for handling multi-font documents:
\begin{description}
\item [Font abstraction.]
On multi-font documents using the Latin script, a high accuracy can potentially be obtained within a limited range of fonts known to an OCR system. As an effort to overcome the difficulties in recognition of documents with a variety of fonts, Kahan et al.~\cite{KahanPAMI1987RecPrtChrAnyFnt} introduced an OCR system utilizing an \emph{abstraction} of the font that generalizes the individual differences of various fonts. This approach is called an \emph{omni-font} optical character recognition, and it has a number or successful implementations.
\item [Font identification.]
As opposed to the font abstraction -- Jung et al.~\cite{JungICDAR1999MltFntClsTypAlg} argue -- font \emph{identification} can give details on the structural and the typographical design of characters. Furthermore, with the font information it is possible to make the OCR system handle a document with a confined effort for an identified font. In other words, an OCR system consisting of a various mono-font segmentation tools and recognizers can perform a font-specific recognition. Their approach uses typographical attributes such as ascenders, descenders and serifs obtained from a word image. The attributes are used as an input to a neural network classifier to produce the multi-font classification results. They claim the method can classify 7 commonly used fonts for all point sizes from 7 to 18. Their experiments have shown font classification accuracy figures reach high levels of about 95\% even with severely touching characters.
\end{description}

Even within a single font there can be variations (distortions) in character print presenting difficulties to OCR systems. Such variations include the use (and sometimes abuse) of \textbf{boldface}, \textit{italics}, \underline{underlining}, or \textsc{small-caps} character styles (typefaces). Some OCR systems attempt to remedy the typeface differences via image pre-processing, while others simply treat them as different fonts.

\subsection{Handwriting recognition}
\begin{figure}[h]
\begin{center}
\begin{tabular}{|c|c|}
\hline
\includegraphics[width=2.8in,height=5in]{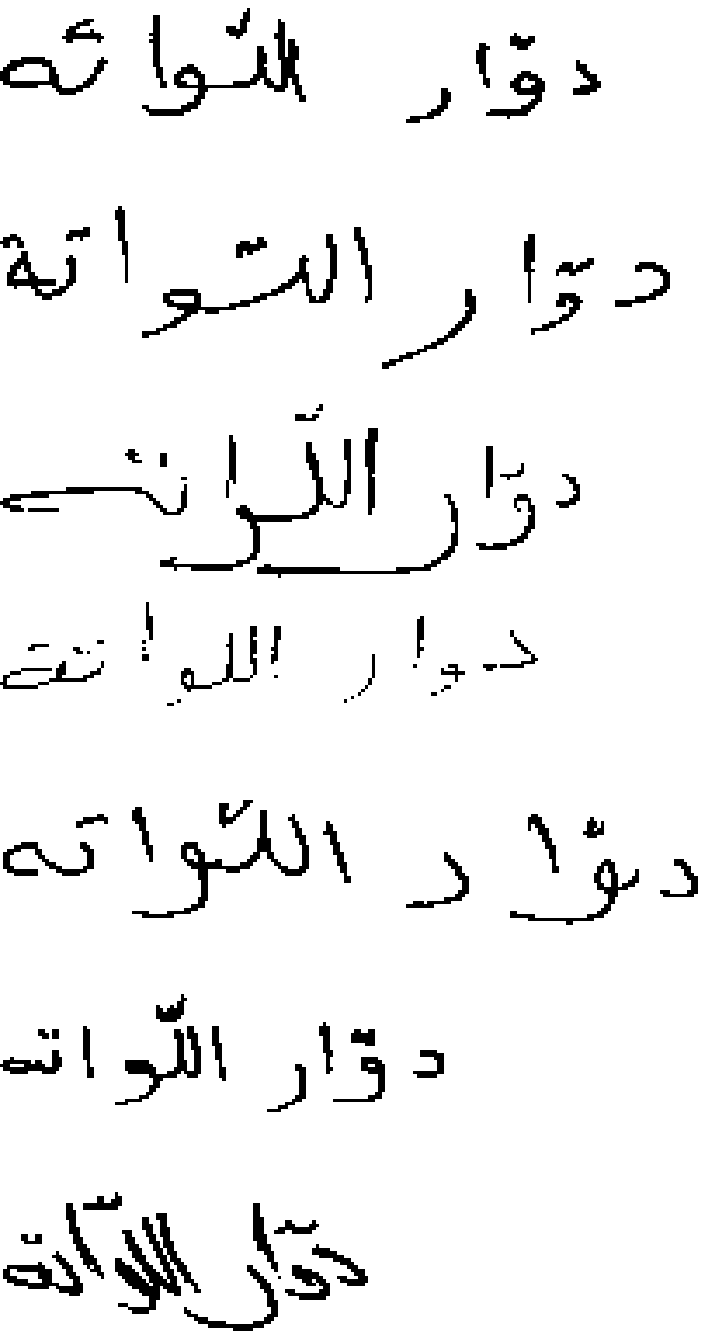} &
\includegraphics[width=2.8in,height=5in]{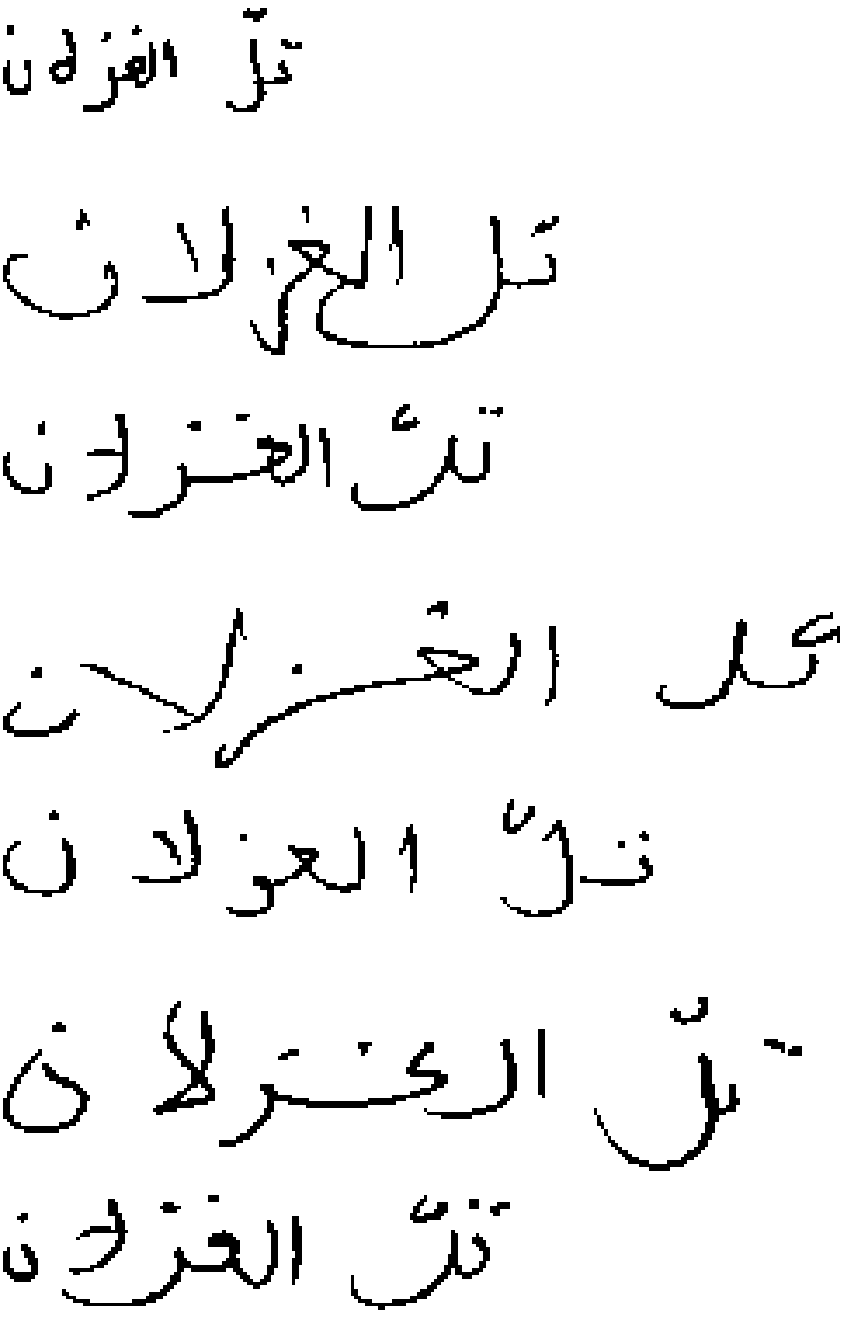} \\
\hline
\end{tabular}
\caption{Names of two Tunisian villages written in Arabic by seven different subjects}
\label{figArabicHandwrittenSamples}
\end{center}
\end{figure}

Recognition of handwritten text, or Intelligent Character Recognition (ICR), is a considerably more difficult task than recognition of typed or machine-printed text. As Edelman et al.~\cite{EdelmanEtal90} point out, the two main reasons for the difficulty are
\begin{itemize}
\item character segmentation ambiguity and 
\item character shape variability
\end{itemize}
See Figure~\ref{figArabicHandwrittenSamples} for examples of Arabic handwriting taken from the corpus described by Pechwitz et al. in~\cite{IFN-ENIT-corpus}. Handwritten text is often cursive, meaning that a considerable part of a word image may consist of ligatures that connect consecutive characters. These ligatures may often be interpreted as parts of letters giving rise to multiple equally plausible segmentations. Furthermore, even when written by the same person, the same characters may change not only geometry (e.g. being slightly slanted in some cases and straight in others) but also topology (e.g. the numeral ``2'' written both with and without a loop at the bottom). Therefore recognition methods are often tuned to a specific handwritten style and have to be retrained or even reprogrammed when a different style is presented.

The core of a complete ICR system is its recognition unit. According to Steinherz et al.~\cite{Steinherz1999CrsvScrWrdRec}, one way of classifying recognition algorithms is by the size and nature of the \emph{lexicon} involved, as well as by whether or not a \emph{segmentation} stage is present. Lexicons may be 
\begin{itemize}
\item small and specific containing about $100$ words,
\item limited, but dynamic, which may go up to $1000$ words, and
\item large or unlimited.
\end{itemize}

Segmentation-based methods take a sequence of primitive segments derived from a word image together with a group of possible words, and attempt to concatenate individual segments to best match candidate characters. In the segmentation-free case, no attempt is made to separate the image into pieces that relate to characters, although splitting into smaller pieces is still possible, followed by generation of a sequence of observations. In both cases, algorithms try to reproduce how a word was {\em written} by recognizing its components in the left-to-right (or right-to-left) fashion.

Alternatively, there is a class of \emph{perception-oriented} methods that perform a human-like {\em reading} by utilizing some stable features to reliably find characters anywhere in the image, and to use them to bootstrap a few word candidates for a final evaluation phase. While this class of methods has not been developed as much as the other two classes, there exist some perception-oriented approaches that appear quite promising.

To date, most attention appears to have been directed toward small- and limited-lexicon methods, especially those applied to Latin scripts. Many different methods and a few complete recognition systems have been proposed in the literature. Overall, however, the field of handwritten text recognition has not yet matured~\cite{Steinherz1999CrsvScrWrdRec}.

There are relatively few review-style articles~\cite{PlamondonSrihari,KhorsheedPAA2002OflACRRv} and even fewer attempts to experimentally compare different methods. One reason for this is lack of sufficiently large, general, and publicly available test datasets.

Furthermore, the performance of a complete ICR system depends on many other components besides its recognition engine, e.g. preprocessing, segmentation, and post-processing modules. To our knowledge, there have been virtually no attempts to create a common framework that allows an objective comparison of different engines or complete recognition systems. The only exception we have found is HRE API, which stands for the \emph{handwriting recognition engine application program interface}. The API was released around 1994 to the Internet by Sun Microsystems in hopes that it would speed the availability of commercial quality handwriting recognition engines and applications, and help foster university research on new handwriting recognition technology. Being an open source fully extensible framework for handwriting recognition, HRE API aimed at:
\begin{itemize}
\item providing a functionally complete interface to handwriting recognition engines,
\item simultaneous support of multiple handwriting recognition engines,
\item minimizing dependence on a specific graphical user interface (GUI), and
\item full multi-language support.
\end{itemize}

The most successful approaches to hand-written character and word recognition are based on the application of \emph{Hidden Markov Models} (HMM). Such approaches model variations in printing and cursive writing as an underlying probabilistic structure, which cannot be observed directly \cite{CaseyLecolinet}. When small lexicons are involved, elementary HMMs that model individual characters are combined to form individual larger {\em model-discriminant} HMMs for every word in the lexicon. When a word image is presented to such a system for recognition, it chooses the word interpretation whose model gives the highest probability given a sequence of observations, or symbols, generated from the given image. For larger lexicons only one {\em path-discriminant} HMM is built from the character sub-HMMs and recognition is performed by finding the most likely path through the model given the sequence of observations from a word image. Path-discriminant models are more general but less accurate than model-discriminant ones.

HMM-based methods are popular because they have performed quite well on the similar problem of speech recognition, and because their mathematical machinery has been developed and refined quite well. At the same time, they still constitute an active area of research. One reason is that performance of an HMM depends strongly on the type of features used to generate observation sequences both for training of the models and for recognition. As pointed out in \cite{Steinherz1999CrsvScrWrdRec}, the ``ideal'' feature set has not yet been found, and may not exist at all. Also, while most HMM-based methods represent a word image as a one-dimensional sequence of observation vectors, a more recent approach is to generalize to two-dimensional Markov Models (also called Markov Random Fields) \cite{ParkLee}. While this approach appears to give better accuracy, it is also much more computationally expensive and so requires more sophisticated design and implementation than traditional HMM-based methods.

\subsection{Flexibility}
The early optical character recognition systems were quite inflexible. They often worked only with a single font or a strictly limited set of fonts (e.g. OCR-A or OCR-B as in the Figures~\ref{figOCRAFont} or \ref{figOCRBFont}), required a fixed document layout, had very limited functional extension options and low tolerance to any noise present in the input documents.

Nowadays, a good optical character recognition system is expected to be flexible with respect to the user's data and tasks. We have identified the three common areas of OCR system flexibility:
\begin{itemize}
\item \emph{task-adjustable configuration parameters} such as recognition quality and speed, expected scripts and font parameters, and unrecognized character marks
\item \emph{trainability} to optimize the recognition engine for a specific document set
\item \emph{software development kit} (SDK) to extend the OCR engine's functionality and/or making it part of a more complex document recognition system
\end{itemize}

Let us consider several examples of the existing OCR solutions mentioned in Section~\ref{secAvlOCRSys}. We shall characterize the flexibility of each solution with respect to the above areas.
\begin{description}
\item [ScanSoft Capture Development System 12] is a flexible OCR SDK providing adjustable throughput/accuracy OCR (machine print) and ICR (hand-print) engines supporting character training to achieve improved accuracy.
\item [ABBYY FineReader 6.0 Engine] is another flexible SDK for integrating ABBYY OCR and ICR recognition technologies into Windows applications. To increase the recognition accuracy of forms and non-standard documents, the engine provides an API for creating and editing special recognition languages, creating user languages as a copy of predefined recognition languages, and adding new words to user languages. The FineReader engine allows developers to create and use user patterns or import them from ABBYY FineReader desktop application.
\item [Sakhr Automatic Reader 7.0] OCR engine is primarily used for Arabic machine print recognition but has multi-language OCR support through Xerox Text Bridge technology. It has a rich set of configurable options and can distinguish between 26 Arabic TrueType fonts. This trainable OCR engine comes with an OCR SDK in both DLL and ActiveX formats.
\end{description}

We should mention some other OCR system flexibility features that are seldom available but often quite desirable:
\begin{itemize}
\item source code availability
\item portability across computing platforms
\item gray-scale and color image support
\end{itemize}
One may find some of those features in open-source OCR systems also mentioned in Section~\ref{secAvlOCRSys}.

\subsection{Accuracy}
Computer based optical character recognition (OCR) systems have gone a long way since their introduction in the early 1950's. Although their productivity, robustness and accuracy have improved much since then, the OCR systems are still far from being perfect especially regarding their \emph{accuracy}.

Given a paper document, one can scan it to produce a digital image that can be processed by an OCR program, whose output is the document's text presented in a machine-readable form, e.g. a text file. To estimate the OCR accuracy, the OCR output can then be compared to the text of the original document, called the \emph{ground truth}. Since there are multiple ways to compare the OCR output to the ground truth, we have multiple ways to define the OCR accuracy.

When the original document's text is available in an electronic format, a computer can evaluate the OCR accuracy by comparing the document's source file with the OCR output file. The accuracy expression usually takes into account the number of
\begin{itemize}
\item correct,
\item inserted,
\item deleted, and
\item substituted
\end{itemize}
items (e.g. words or characters) in the OCR output with respect to the source text. Refer to Section~\ref{secGrTrPr} for details on the ground truth based accuracy estimation.

When the document's source text is not available, a fluent native speaker of the language can type the correct text into a file, in order to provide a ground truth. Alternatively, one resorts to having a native speaker look and visually compare the source document and the  corresponding OCR output. This process may be as casual as identifying misspelled and missing words in the OCR output, or it can follow a rigorous path of computing a formal OCR accuracy score. In Section~\ref{secGrTrAb} we give details of approximate accuracy evaluation without the document's source text.

\subsubsection{When ground truth is present\label{secGrTrPr}}
The two most common OCR accuracy measures are \emph{precision} and \emph{recall}. Both are \emph{relative} measures of the OCR accuracy because they are computed as ratios of the correct output to the total output (precision) or input (recall). More formally defined,
\begin{eqnarray}
precision=\frac
{\textit{number of correct items}}
{\textit{number of items in OCR output}} \\
recall=\frac
{\textit{number of correct items}}
{\textit{number of items in ground truth}}
\end{eqnarray}
where \textit{items} refer to either characters or words, and \textit{ground truth} is the original text stored as e.g. a plain text file. In calculating \emph{precision} and \emph{recall}, the ground truth is used to
\begin{itemize}
\item determine the number of correctly recognized items and
\item supply the total number of the ground truth items.
\end{itemize}
Both \emph{precision} and \emph{recall} are mathematically convenient measures because their numeric values are some decimal fractions in the range between 0.0 and 1.0, and thus can be written as percentages. For instance, \emph{recall} is the percentage of words in the original text correctly found by the OCR engine, whereas \emph{precision} is the percentage of correctly found words with respect to the total word count of the OCR output. Note that in the OCR-related literature, the term OCR \emph{accuracy} often refers to \emph{recall}.

Consider a simple example. Suppose we receive a one page fax that contains mostly text. We run our favorite OCR program to convert it to a plain text file \texttt{output.txt}, our OCR output. We then receive the original text file \texttt{input.txt} (our ground truth) via e-mail. Having both files at our disposal, we discover that \texttt{input.txt} has 3371 characters in 544 words, while \texttt{output.txt} contains 3374 characters in 542 words. Evaluation reveals that \texttt{output.txt} has 3350 correctly matched characters and 529 correctly matched words. The precision and recall values are then computed as shown in Table~\ref{tblExClcPR}.
\begin{table*}[h]
\centering
\begin{tabular}{c||c|c||c|c}
 & \multicolumn{2}{c||}{character-wise} & \multicolumn{2}{c}{word-wise} \\
\cline{2-5} 
measure & calculation & percentage & calculation & percentage \\
\hline
\emph{recall} & $\frac{3350}{3371}=0.994$ & 99.4\% & $\frac{529}{544}=0.972$ & 97.2\% \\
\hline
\emph{precision} & $\frac{3350}{3374}=0.993$ & 99.3\% & $\frac{529}{542}=0.976$ & 97.6\% \\
\end{tabular}
\caption{An example of precision and recall calculation}
\label{tblExClcPR}
\end{table*}

While precision and recall are good OCR performance measures, they remain two strongly correlated but nevertheless separate quantities. OCR software users often require a single accuracy measure formulated as a character or word \emph{error rate}.

The reason for using error metrics to measure system performance is that error metrics represent the cost to the user of having the system make those errors. Cutting the error rate by a factor of two, for example, is an indication that the cost to the user is also cut in half in that, if the user were to correct those errors, one would have to devote only half as much effort. Improvements in system performance can then be tracked by measuring the relative decrease in error rate.

Makhoul et al.~\cite{MakhoulDBNW1999PrfMsrInfExt} present precision and recall in terms of item \emph{insertion}, \emph{deletion} and \emph{substitution} errors, and analyze several approaches to OCR error rate estimation.

The classical \emph{effectiveness measure}~\cite[p.174]{vanRijsbergen1979IR} is given by
\begin{equation}
M=1-\left[\frac{\alpha}{P}+\frac{(1-\alpha)}{R}\right]^{-1}
\end{equation}
where $P$ is precision, $R$ is recall, $\alpha \in [0,1]$. Makhoul shows that $M$ underweighs the role of \emph{insertion} and \emph{deletion} errors with respect to the \emph{substitution} errors, and therefore presents OCR performance as better than it actually is. To avoid this problem Makhoul proposes the \emph{item error rate} formula:
\begin{equation}
E=\frac
{insertions+deletions+substitutions}
{\textit{number of items in ground truth}}
\label{eqTotalError}
\end{equation}
Having a denominator that is fixed for a given test document, this error rate estimate provides a good way to compare performances of different OCR systems. Note that the item error rate is analogous to the word error rate used as the primary performance measure in speech recognition systems, but it is not perfect either: its numeric range is not restricted between 0.0 and 1.0 primarily because it does not take into account the OCR output size.

MacKenzie and Soukoreff~\cite{MacKenzieNordiCHI2002ChLvlErrTxtEnt} suggest a more fair item error rate estimate
\begin{equation}
E=\frac
{insertions+deletions+substitutions}
{max(\textit{number of items in ground truth},\textit{number of items in OCR output})}
\end{equation}
This expression for the error rate is unbiased with respect to insertions, deletions and substitutions, its numerical range is restricted between 0.0 and 1.0, and it accounts for both the ground truth size and the OCR output size. The only evident drawback is that for a given input document the denominator in the error expression is not fixed across different OCR systems, and this fact complicates the fair comparison of their performances.

Nartker et al.~\cite{NartkerSPIE2003OCRCrcDcmLvlKnw} attempt to refine the notion of OCR accuracy as being applied for information retrieval (IR) tasks. They define
\begin{equation}
CharacterAccuracy=\frac
{TotalCharacters-CharacterErrors}
{TotalCharacters}
\label{eqCharAcc}
\end{equation}
where $TotalCharacters$ is the number of characters in the ground truth, and $CharacterErrors$ is the total number of character insertions, deletions and substitutions. Notice that such defined $CharacterAccuracy$ can be negative, which can confuse the interpretation of accuracy, but will be consistent with eq.(\ref{eqTotalError}) if we write $CharacterAccuracy=1-E$.

Nartker also notes that IR systems ignore commonly occurring words, called \emph{stopwords} (such as ``the", ``or", ``and"). They also ignore punctuation, most numeric digits, and all stray marks. Thus, misspelled stopwords and incorrect numbers are not errors that affect retrieval performance. In terms of retrievability of documents from an IR system, \emph{non-stopword accuracy} is a better measure of conversion accuracy. Non-stopword accuracy is defined as follows:
\begin{equation}
NonStopwordAccuracy=\frac
{TotalNonStopwords-NonStopwordErrors}
{TotalNonStopwords}
\end{equation}
Here again $TotalNonStopwords$ are counted in the ground truth, and $NonStopwordErrors$ only account for those errors that occur in non-stopwords. Notice that as in eq.(\ref{eqCharAcc}), the accuracy estimate here can be negative.

\subsubsection{When ground truth is absent\label{secGrTrAb}}
There are many practical occasions when the source text is not available and thus no ground truth exist to verify the OCR output automatically. In those cases we resort to evaluating the performance of an OCR engine by a fluent native speaker, considering the human perceptual system to be the ultimate optical character recognition system. Needless to say, all these performance evaluations would be approximate and biased. Below we provide some guidelines on what to do when we do not have a set of text files to compare the OCR output to.
\begin{enumerate}
\item \label{lbA1Begin}Have a fluent speaker of the language examine a source page image and the OCR output.
\item Have the speaker identify words in the OCR output that are wrong, missing, and extra.
\item \label{lbA1End}Estimate the number of words on the page and use a formula from Section~\ref{secGrTrPr}.
\item Repeat steps~\ref{lbA1Begin}-\ref{lbA1End} for a selected variety of pages.
\end{enumerate}
For example, if the page has about 300 words, and the speaker identifies that a paragraph of about 20 words is missing, 3 words are incorrect, and there are no extra words, then word-level precision for the page is about 99\%, and word-level recall for the page is about 92\%.

The above OCR evaluation method should be applied to document images with various kinds of noise (e.g. due to copying or faxing), as well as to some clean images. This way, one can get a sense of what kinds of OCR accuracy numbers appear in different situations.

When great volumes of OCR data are to be verified, the verification process can be automated and the method described above is applicable here as well, if in place of a fluent speaker one considers a computer equipped with an alternative OCR system. When several OCR engines are available, one can match their output, thereby detecting possible errors, and suggest possible corrections based on statistical information and dictionaries~\cite{KleinIROCR2002VotSys4AutOCRCrc}.

\subsubsection{Accuracy of commercial OCR engines}
Summers~\cite{Summers2001OCRAcr3Sys} compared the performance of three leading commercial OCR packages on document images of varying quality. The OCR packages included
\begin{itemize}
\item Caere Developers Kit 2000 version 8 by ScanSoft,
\item FineReader 4.0 by Abbyy, and
\item NeuroTalker 4.1 from International Neural Machines.
\end{itemize}
The precision and recall of each system was measured on the English corpora that consisted of 1147 journal page images from the University of Washington corpus, 946 page images from a set of declassified government documents, and 540 page images from  tobacco litigation documents.
\begin{table*}[h]
\centering
\begin{tabular}{|c|cc|cc|cc|}
\hline
 & \multicolumn{2}{c|}{CDK 2000} & \multicolumn{2}{c|}{FineReader 4.0} & \multicolumn{2}{c|}{NeuroTalker 4.1} \\
corpus & precision & recall & precision & recall & precision & recall \\
\hline
Tobacco & 91.65\% & 91.73\% & 80.98\% & 87.61\% & 80.80\% & 80.51\% \\
U.Washington & 98.53\% & 74.75\% & 97.73\% & 74.66\% & 94.85\% & 72.81\% \\
Declassified & 76.99\% & 75.58\% & 74.86\% & 80.64\% & 71.54\% & 61.94\% \\
\hline
total: & 89.38\% & 78.53\% & 86.08\% & 79.47\% & 78.60\% & 70.49\% \\
\hline
\end{tabular}
\caption{Document average results}
\label{tblOCRAccrDocAvgRes}
\end{table*}
Among other statistics, \emph{document average} values for precision and recall were computed. As Table~\ref{tblOCRAccrDocAvgRes} shows, Caere Developers Kit (CDK) 2000 and FineReader 4.0 consistently outperformed NeuroTalker, and in general CDK 2000 was slightly more accurate than FineReader 4.0. In the case of the declassified government documents, however, FineReader 4.0 exhibited somewhat superior recall to that of CDK 2000.

Kanungo et al.~\cite{Kanungo1999OmniPageSakhr} propose a statistical technique called the \emph{paired model} to compare the accuracy scores of two recognition systems: Sakhr Automatic Reader 3.0 and Shonut OmniPage 2.0. They show that the paired model approach to performance comparison gives rise to tighter confidence intervals than unpaired methods. The method was tested on the DARPA/SAIC dataset consisting of 342 documents of Arabic text with the original resolution of 600 dpi, artificially sub-sampled to 400, 300, 200, and 100 dpi. Their experiments indicate that on the 300 dpi SAIC dataset Sakhr has higher \emph{accuracy} than OmniPage but OmniPage has a better \emph{precision}. They have noticed that the Sakhr accuracy drops when the image resolution is increased beyond 300 dpi. Furthermore, they have found that the average time taken for Sakhr to OCR a page does not increase when the image resolution is increased from 400 dpi to 600 dpi. In their view, this may indicate that Sakhr internally converts input images to some standard resolution prior to the recognition stage.

\subsection{Productivity}
Along with the recognition accuracy, \emph{productivity} (or \emph{throughput}) is another important property of an OCR system. The throughput is measured in the number of items processed per an interval of time, e.g. pages per second or words per minute. In general, the throughput of an OCR system is affected by a number of factors that include
\begin{itemize}
\item \emph{character classifier} employed by the system
\item \emph{document complexity} determined by e.g. the number of fonts, text columns, non-text items (pictures, plots, drawings) on each page
\item \emph{image noise} due to e.g. a poor quality original or imperfect scanning
\item \emph{post-processing} often required for OCR correction and robust document indexing
\item \emph{interface issues} involving input image pre-processing and output text formatting
\item \emph{handwriting} recognition option
\end{itemize}

Not surprisingly, task-specific OCR systems are usually more accurate and productive than their general-purpose counterparts, but the latter are expected to handle a greater variety of documents. Below, we give some examples of both kinds of OCR systems and indicate their typical \emph{average throughput}.

\subsubsection{General purpose readers}
Since general purpose OCR systems are designed to handle a broad range of documents, some sacrifices in recognition accuracy and average throughput are expected. Some high-end OCR packages allow the users to adapt the recognition engine to customer data to improve recognition accuracy. They also can detect type faces (such as bold face and italic) and output the formatted text in the corresponding style. This versatility often comes at the expense of the system's productivity.

\begin{description}
\item [PrimeRecognition] ({\tt http://primerecognition.com/}) claims that on a 700MHz Pentium PC their OCR engine's throughput is 420 characters/sec. with the average OCR accuracy rate of 98\%.
\item [LeadTools] ({\tt http://www.leadtools.com/}) intelligent character recognition (ICR) engines on a 1500MHz Pentium PC with 256MB of RAM running Windows 2000 are claimed to achieve recognition speeds of 280-310 characters per second for hand-printed numerals (alone) and 150-220 characters per second for hand-printed upper and lower case text, numbers and symbols (intermixed).
\item [Sakhr] ({\tt http://www.sakhr.com/}) produces Automatic Reader primarily used for Arabic OCR. Aramedia ({\tt http://aramedia.net/aramedia/ocr.htm}) claims that Automatic Reader 7.0 running on Pentium III under Windows 2000 can process
800 characters per second with 99\% accuracy in recognizing Arabic books and newspapers.
\end{description}

\subsubsection{Task-specific readers}
As was mentioned before, task-specific readers are used primarily for high-volume applications targeting similarly structured documents and requiring high system throughput. To help increase the throughput, such systems focus on the document's \emph{fields of interest}, where the desired information is expected to be. This narrow-focused approach can considerably reduce the image processing and text recognition time. Some application areas where task-specific readers are employed include: 
\begin{itemize}
\item ZIP code recognition and address validation
\item form-based text recognition, e.g. in tax forms
\item automatic accounting procedures for paper utility bills
\item bank check processing
\item accounting of airline passenger tickets
\item signature verification
\end{itemize}
To be more specific, let us consider some examples of task-specific OCR systems.
\begin{description}
\item [Address readers] help automatically sort mail. A typical address reader first locates the destination address block on a mail piece and then reads the ZIP code in this address block. If additional fields in the address block are read with high confidence the system may generate a 9 digit ZIP Code for the piece. The resulting ZIP code is used to generate a bar code subsequently printed on the envelope for automatic mail sorting. A good example of an address reader is Multiline Optical Character Reader (MLOCR) used by the United States Postal Service (USPS). It can recognize up to 400 fonts and the system can process up to 45,000 mail pieces per hour.
\item [Form readers] help automate data entry by reading data in paper forms. A typical form reading system should be able to discriminate between pre-printed form instructions and filled-in data. Such a system is first trained with a blank form to register those areas on the form where the data should appear. During the recognition phase, the system is expected to locate and scan the regions that should be filled with data. Some readers read hand-printed data as well as various machine written text. Typically, form readers can process forms at a rate of 5,800 forms per hour.
\item [Airline ticket readers] are used nowadays by several airlines in passenger revenue accounting systems to accurately account for passenger revenues. In order to claim revenue from a passenger ticket, an airline needs to have three records matched: reservation record, the travel agent record and the passenger ticket. The OCR-assisted system reads the ticket number on a passenger ticket and matches it with the one in the airline reservation database. A typical throughput of such a system is 260,000 tickets per day with a sorting rate of 17 tickets per second.
\end{description}

\section{OCR pre-processing}\label{secOCRpreproc}
Modern OCR engines are expected to handle a variety of document imagery that can have wide variations in image quality, noise levels, and page layout. Thus, some document image filtering and page structure analysis become important prerequisites for the actual recognition stage. This section sheds some light on a variety of document image defects, exposes their potential causes, and discusses some technological approaches to enhance document imagery for OCR. Here we also touch some interesting issues pertaining to pre-OCR automatic page and document segmentation.

\subsection{Document image defects}
More often than not, real document imagery can be imperfect. Various document image defects such as 
\begin{itemize}
\item white or black speckle,
\item broken or touching characters,
\item heavy or light print,
\item stray marks,
\item curved baselines
\end{itemize}
occur during document printing/scanning and have many physical causes including
\begin{itemize}
\item dirty originals, e.g. from coffee stains
\item paper surface defects
\item spreading and flaking of ink or toner
\item optical and mechanical deformation and vibration
\item low print contrast
\item non-uniform illumination
\item defocusing
\item finite spatial sampling resolution
\item variations in pixel sensor sensitivity and placement
\item noise in electronic components
\item binarization or fixed and adaptive thresholding
\item copy generation noise accumulation from previous scanning and copying
\end{itemize}
The accuracy of document recognition systems falls quite abruptly even with slight input document image degradation. To overcome these difficulties, researchers continue to study models of image defects to help control the effects of variation in image quality and construct classifiers meeting given accuracy goals. Such models are often used in synthesizing large realistic datasets for OCR training.

Baird~\cite{BairdIAPR1993DocImgDfcMdl} discusses models of document image defects. He distinguishes two generic approaches to specifying such models: \emph{explanatory} and \emph{descriptive}. Explanatory models, being based on physics, can be validated in part by referencing laws of physics. This can lead to accurate models, but they may be unnecessarily specific and complicated. Descriptive models, being more empirical, are validated principally by statistical measures, e.g. the probability of generating duplicates of real defective image. In general, researchers agree that descriptive models are more practical.

Sarkar et al.~\cite{SarkarICDAR2003TrnSvrDgr} use a ten-parameter image degradation model to approximate some aspects of the known physics of machine-printing and imaging of text, including symbol size, spatial sampling rate and error, affine
spatial deformations, speckle noise, blurring, and thresholding. They  show that document image decoding (DID) supervised training algorithms can achieve high accuracy with low manual effort even in the case of severe image degradation in both training and test data. The authors describe improvements in DID training of character template, set-width, and channel (noise) models. Their large-scale experiments with synthetically degraded images of text showed that the method can achieve 99\% recall even on severely degraded images from the same distribution. This ability to train reliably on low-quality images that suffer from massive fragmentation and merging of characters, without the need for manual segmentation and labeling of character images, significantly reduces the manual effort of DID training.

Image warping is a common problem when one scans or photocopies a document page from a thick bound volume, resulting in shading and curved text lines in the spine area of the bound volume. Not only does this impair readability, but it also reduces the OCR accuracy. To address this problem Zhang and Tan~\cite{ZhangICDAR2003CrctDocImgWrp} propose a page image un-warping technique based on connected component analysis and regression. The system is claimed to be computationally efficient and resolution independent. They report that the overall average precision and recall on 300 document images are improved by 13.6\% and 14.4\% respectively.

\subsection{Document image enhancement}
OCR performance places an upper limit on the usefulness of any text-based processing of scanned document images. Document image quality, in turn, can greatly affect the OCR engine accuracy and productivity. While dealing with large collections of marginal quality document imagery, one usually faces a problem of automatic document image enhancement and restoration for OCR.

Loce and Dougherty~\cite{LoceDougherty1997} provide an interesting treatment of digital document enhancement and restoration problem via non-linear image filtering algorithms and the statistical optimization techniques behind them. In particular, the book describes optimization methods for parallel \emph{thinning}, \emph{thickening}, and \emph{differencing} filters. The statistical optimality of these filters is described relative to minimizing the mean-absolute error of the processed document with respect to the ideal form of the document. The authors describe how to use \emph{increasing filters} in restoration methods for degraded document images that exhibit noticeable background noise, excessively thick character strokes, and ragged edges. The book also discusses several widely used techniques for \emph{spatial resolution} and \emph{quantization range} enhancement.

Let us discuss some of the noteworthy papers on OCR-related image enhancement.
Cannon et al.~\cite{Cannon1999quarc}
introduced QUARC, a system for enhancing and restoring images of typewriter-prepared documents. They devised a set of quality assessment measures and image restoration techniques to \emph{automatically} enhance typewritten document imagery for OCR processing. They used five such quality measures:
\begin{itemize}
\item small speckle factor
\item white speckle factor
\item touching character factor
\item broken character factor
\item font size factor
\end{itemize}
The quality measures were used to come up with an optimal restoration method for a given document image. Such restoration methods were based on the following bi-tonal filters:
\begin{itemize}
\item do nothing
\item cut on typewriter grid
\item fill holes and breaks
\item despeckle
\item global morphological close
\item global fill holes and breaks
\item global despeckle
\item kFill filter
\end{itemize}
The method characterized an input document image via the quality measures, whose values were used by a pre-trained \emph{linear classifier} to determine an optimal image restoration filter. The employed classifier was trained on a set of ``typical" images whose ``best" restoration methods were determined by the performance of a given OCR engine. QUARC was claimed to have improved the character error rate by $38$\% and the word error rate by $24$\%. The system assumed input documents in fixed typewriter font of Latin script, and did not address documents with font and script variations.

Summers et al. have employed a similar approach in the document image enhancing software called \emph{ImageRefiner}~\cite{Summers2003ImageRefiner}. They improved upon QUARC by addressing documents in non-fixed fonts (e.g. TrueType) and non-Latin (e.g. Cyrillic) scripts. The QUARC quality measure set was extended by adding the following measures of \emph{connected components}: size, height, width, area, and aspect ratio.
while adapting the set of restoration methods that did not assume a \texttt{typewriter} grid. Similarly to QUARC, their system determined an optimal restoration method based on the document image quality measures, but their classifier employed an \emph{error propagation neural net} to learn an optimal classification mapping between the feature space (quality measures) and the document classes (restoration methods). ImageRefiner was robust while working on noisy document images with various fonts and scripts. It proved to be able to pick the correct transformation 76.6\% of the time. Currently, they are working on the second phase of ImageRefiner that is expected to use adaptive document zoning and handle non-Latin-like (e.g. Asian) scripts.

\subsection{Document segmentation}
In order to ensure optimal character recognition, searchability and minimum distortion, there is often the need to preserve a document's logical structure and format when it is converted into electronic form. One would ideally like a document recognition system to behave in a ``what you scan is what you get" fashion. This has stimulated research in an area often referred to as Document Image Understanding. It encompasses
\begin{itemize}
\item decomposing a document into its structural and logical units (text blocks, figures, tables, etc.),
\item tagging these units,
\item determining their reading order,
\end{itemize}
and thus ultimately recovering the document's structure. In addition, format analysis, which includes the analysis of font types and spacing, has to be performed. This section surveys methods of automatic document page segmentation, recovering reading order and finding text lines.

\subsubsection{Text versus other material} 
Mukherjee and Acton~\cite{Mukherjee1999DocPgSgmMltScl} describe a \emph{multi-scale clustering} technique for document page segmentation. Unlike the hierarchical multi-resolution methods, this document image segmentation technique simultaneously uses information from different scaled representations of the original image. The final clustering of image segments is achieved through a fuzzy c-means based similarity measure between vectors in scale space. The segmentation process is claimed to reduce the effects of insignificant detail and noise while preserving the objects' integrity. Their results indicate that the scale space approach allows meaningful page segmentation effective for document coding. One of the drawbacks is the additional cost incurred in generating the scale space itself. The authors are currently optimizing their method for computational speed and are developing object based coding algorithms that complement the segmentation approach.

Andersen and Zhang~\cite{AndersenICDAR2003FtrNNRgnId} present several critical features that were used in a \emph{neural network}-based region identification algorithm applied to newspaper document segmentation. Their approach is claimed to achieve good identification accuracy (98.09\%), especially surprising in light of the degree to which the regions are over-segmented. The authors indicate a possibility that the segmentation results could be improved even further, by implementing a soft decision neural net with two output nodes (one node representing text and the other node representing non-text) to classify some regions as non-homogenous, i.e. containing both text and non-text. With this configuration it should then be possible to re-partition this type of regions into homogenous parts, which should open the possibility for the classification pass to correct some of the mistakes from the initial segmentation pass. Their current work is also focusing on improving classification accuracy by using contextual information (information from adjacent regions) as well as features from the gray-scale document. Initial work has shown that this can reduce errors by up to 50\%.

Ma and Doermann~\cite{MaCDRR2003BtStrpStrPgSgm} present an approach to \emph{bootstrapped learning} of a page segmentation model. Their idea originated from attempts to segment document images of a \emph{consistent page structure} (e.g. dictionaries), and was extended to the segmentation of more general structured documents. They argue that in cases of highly regular page structure, the layout can be learned from examples of only a few pages by first training the system using a small number of samples, and processing a larger test set based on the training result. The training phase can be iterated a number of times to refine and stabilize the learning parameters. They have applied this segmentation to many structured documents such as dictionaries, phone books, spoken language transcripts, and have obtained quite satisfying segmentation performance. The authors acknowledge that structured documents containing pictures, figures and tables may be problematic for the current implementation of the method, because pages with such non-text elements usually lack consistent structure.

Zheng et al.~\cite{ZhengCDAS2002SgmIdtHdwrNsyDoc} present an approach to the problem of identifying and \emph{segmenting handwritten annotations} in noisy document images. Such handwritten annotations commonly appear as an authentication mark being (part of) a note, correction, clarification, instruction, or a signature. Their approach consists of two processes:
\begin{itemize}
\item \emph{page segmentation}, dividing the text into regions at the appropriate level: character, word, or zone
\item \emph{classification}, identifying the segmented regions as handwritten
\end{itemize}
To determine the approximate region size where classification can be reliably performed, they conducted experiments at the character, word and zone level. Reliable results were achieved at the word level with a classification accuracy of 97.3\%. Their experiments show that the method may present many false identifications in excessively noisy documents. They address this problem in their subsequent research~\cite{ZhengICDAR2003TxtIdNsyDocImgMRF} by employing a Markov Random Field (MRF) to model the geometrical structure of the printed text, handwriting and noise to rectify misclassifications.

\subsubsection{Finding text lines}
Precise identification of text lines is an important part for most OCR systems. It is also very useful for document layout analysis. Numerous methods have been proposed for text line and baseline finding. Some methods attempt to find just text lines, e.g. by using Hough transforms~\cite{HoughICHEA1959MachAnlsBblPic,SrihariMVA1989AnlsTxtImgHhgTrf}, projection profiles, and Radon transforms. Others find text lines as part of more general document layout analysis tasks, e.g. XY cuts, whitespace segmentation, Voronoi diagrams, and distance-based grouping. Usually, such methods start by performing a rough analysis of the layout, often based on the proximity of bounding boxes of connected components or connected components themselves. More precise base- and text-line models are employed in subsequent steps, as required~\cite{HaralickCVPR1994DocImgUndGmtLgcLyt,OkunSCIA1999RbsDocSkwDet}.

Breuel~\cite{BreuelCDRR2001RbstL2BsLnFdg} presents an algorithm that finds precise page rotation and reliably identifies text lines. The method uses a \emph{branch-and-bound} approach to globally optimal line finding. At the same time it optimizes the baseline and the descender line search under a robust least squares (or maximum likelihood) model. The algorithm is easy to implement and has a very limited number of free parameters (error bound, ranges of permissible page rotations and descender sizes, and numerical accuracy). The method is apparently sensitive to document degradations where large numbers of characters are merged into the same connected component. The algorithm was tested on the English-language memo collection of
the University of Washington Database 2, which consisted of 62 scanned memos, exhibiting a variety of fonts and manual annotations, and presenting a wide variety of image noise. The text line fitting algorithm correctly identified all lines present in the documents when the extracted bounding boxes allowed it to do so. No spurious text lines were detected, and some short one-word lines were ignored. One particularly interesting property of this algorithm is that it allows variations in text line orientations. This permits the algorithm to be applied to document images captured by photo or video cameras rather than by scanners.

Bai and Huo~\cite{BaiICDAR2003ExtTxtLnPenScn} present an approach to extracting the target text line from a document image captured by a hand-held \emph{pen scanner}. Given a binary image, a set of possible text lines are first formed by the nearest-neighbor grouping of connected components. The text line set is then refined by text line merging and adding the missing connected components. A possible target text line is identified via some geometric feature based score function. An OCR engine reads this target line and if the confidence of recognition is high, the target text line is accepted, otherwise, all the remaining text lines are fed to the OCR engine to verify whether an alternative target text line exists or the whole image should be rejected. The effectiveness of the above approach was confirmed by a set of experiments on a testing database consisting of 117 document images captured by C-Pen and ScanEye pen scanners. For C-Pen, out of 113 lines, 112 were detected correctly, 1 wrongly split and none wrongly deleted. For ScanEye, out of 121 lines, 120 were detected correctly, 1 wrongly split and none wrongly deleted.

\subsubsection{Text reading order detection}
Determining the reading order among blocks of text is critically important for OCR systems used in creating and maintaining digital libraries (DL) since it would allow more fully automatic navigation through images of text. This capability also helps OCR systems that utilize high-level context to increase accuracy. 

According to Baird~\cite{BairdICDAR2003DLDIA} automatic reading order detection remains an open problem in general, in that a significant residue of cases cannot be disambiguated automatically through document layout analysis alone, but seem to require linguistic or even semantic analysis.

In a typical scenario when the number of ambiguous cases on one page is rather small, the text ordering ambiguity problem is solved in practice by a judiciously designed interactive GUI presenting ambiguities in a way that invites easy selection or correction. Such capabilities exist in specialized high-throughput scan-and-conversion service bureaus, but are not commonly available in the consumer-level OCR systems.

\begin{figure}
\centering
\epsfig{file=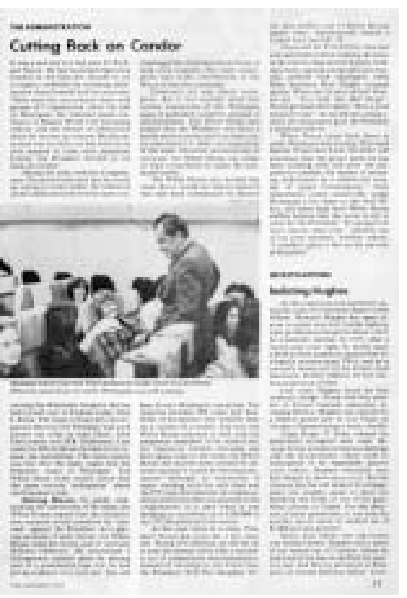}
\epsfig{file=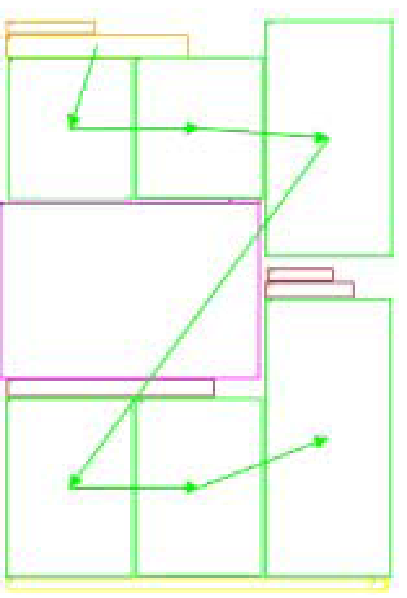}
\epsfig{file=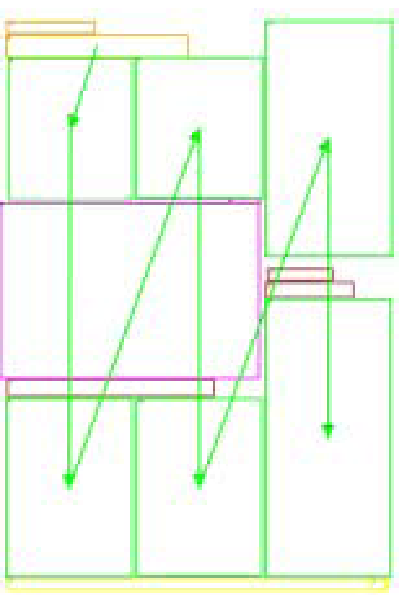}
\caption{An example of document reading order detection}
\label{figRdnOrdDet}
\end{figure}

Aiello et al.~\cite{AielloIJDAR2002DocUdrBrdClsDocs} present a document analysis system that is capable of assigning logical labels and extracting the reading order from a broad set of documents. Their document analysis employs multiple information sources: from geometric features and spatial relations to textual features and content. To deal effectively with these information sources, they define a document representation general and flexible enough to represent complex documents. The possible reading orders are detected independently for document object type \emph{body} and \emph{title}, respectively. Then, these reading orders are combined using a \emph{title-body} connection rule, which connects one \emph{title} with the left-most top-most \emph{body} object, situated below the \emph{title}. The reading order is determined by deploying both geometrical(based on the spatial relations) and content (based on lexical analysis) information. Experiments have been performed using two collections of documents: the University of Washington UW-II English journal database (623 pages), and the MTDB (171 pages) from the University of Oulu, Finland. Figure~\ref{figRdnOrdDet} shows an example of a document and its possible reading orders detected by the spatial analysis alone. The NLP-based lexical analysis module found that the first reading order is the correct one.


\section{OCR post-processing}\label{secOCRpostproc}
While using their ever improving but nevertheless limited recognition abilities on less than perfect real-world document imagery, OCR systems are prone to make mistakes. It is therefore expected that an OCR output will contain errors: OCR noise. Some systems attempt to correct noisy OCR output (e.g. via spell-checkers), some use high-level context (e.g. lexicons) to iteratively improve OCR accuracy, while others simply choose to live with noisy OCR output and account for possible OCR errors during subsequent indexing and text retrieval stages (e.g. via fuzzy string matching). In either case, one has to deal with some kind of OCR post-processing.

Spitz~\cite{SpitzICDAR1995OCRSahpeCode} describes an OCR method which relies on the transformation of the text image into character \emph{shape codes}, a rapid and robust recognition process, and on special lexica which contain information on the "shape" of words and the character ambiguities present within particular word shape classifications. The method currently takes into account the structure of English and the high percentage of singleton mappings between the shape codes and the characters in the words. Considerable ambiguity is removed via lookups in the specially tuned and structured lexicon and substitution on a character-by-character basis. Ambiguity is further reduced by template matching using exemplars derived from surrounding text, taking advantage of the local consistency of font, face and size as well as image quality.

Klein and Kopel~\cite{KleinIROCR2002VotSys4AutOCRCrc} present a post-processing system for enhancing OCR accuracy. They describe how one can match output of several OCR engines, thereby detecting possible errors, and suggest possible corrections based on statistical information and dictionaries. They have tested their system on collections of Hebrew, English and French texts, using two different OCR engines per each language. According to their test results, the proposed system can significantly reduce the error rates, while improving the efficiency of the correction process, making large-scale automatic data acquisition systems closer to becoming a reality.

Beitzel et al.~\cite{Beitzel2002RtrvOCRTxt} give a short survey of OCR text retrieval techniques. They address the OCR data retrieval problem as a search of ``noisy" or error-filled text. The techniques they discuss include: defining an IR model for OCR text collections, using an OCR-error aware processing component as part of a text categorization system, auto-correction of errors introduced by the OCR process, and improving string matching for noisy data.

\section{Summary}
The intent of this report was to review the major trends in the field of modern \emph{optical character recognition} (OCR). We have introduced OCR as a successful branch of \emph{pattern recognition} (a more generic discipline in machine vision), described a typical OCR system and its constituent parts, defined two principle categories of OCR systems (general-purpose and task-specific), outlined the essential OCR techniques, and mentioned many existing OCR solutions (both commercial and public domain).

We have also discussed common OCR difficulties, analyzed their potential causes, and suggested possible ways of resolving them. Discussed in detail were various issues related to 
\begin{itemize}
\item script and language
\item document image types and image defects
\item document segmentation
\item character types
\item OCR flexibility, accuracy and productivity
\item hand-writing and hand-printing
\item OCR pre- and post-processing
\end{itemize}
During the course of our discussion we mentioned the areas of active research as well as the existing OCR systems relevant to each problem. All OCR accuracy and performance figures were taken from the referenced publications. Any additional benchmarking was outside the scope of this survey.

\subsection{Conclusions}
\begin{figure}
\centering
\epsfig{file=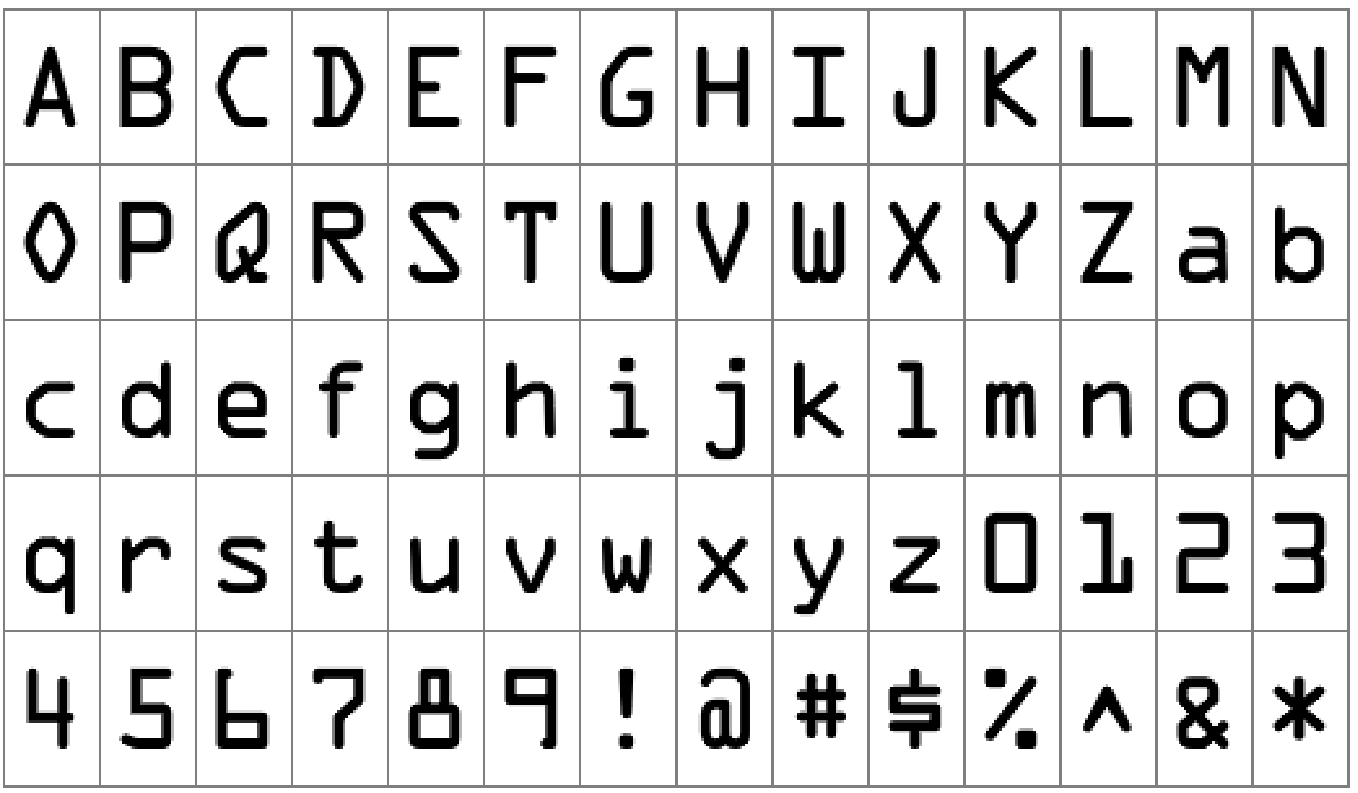}
\caption{OCR-A font}
\label{figOCRAFont}
\end{figure}

\begin{figure}
\centering
\epsfig{file=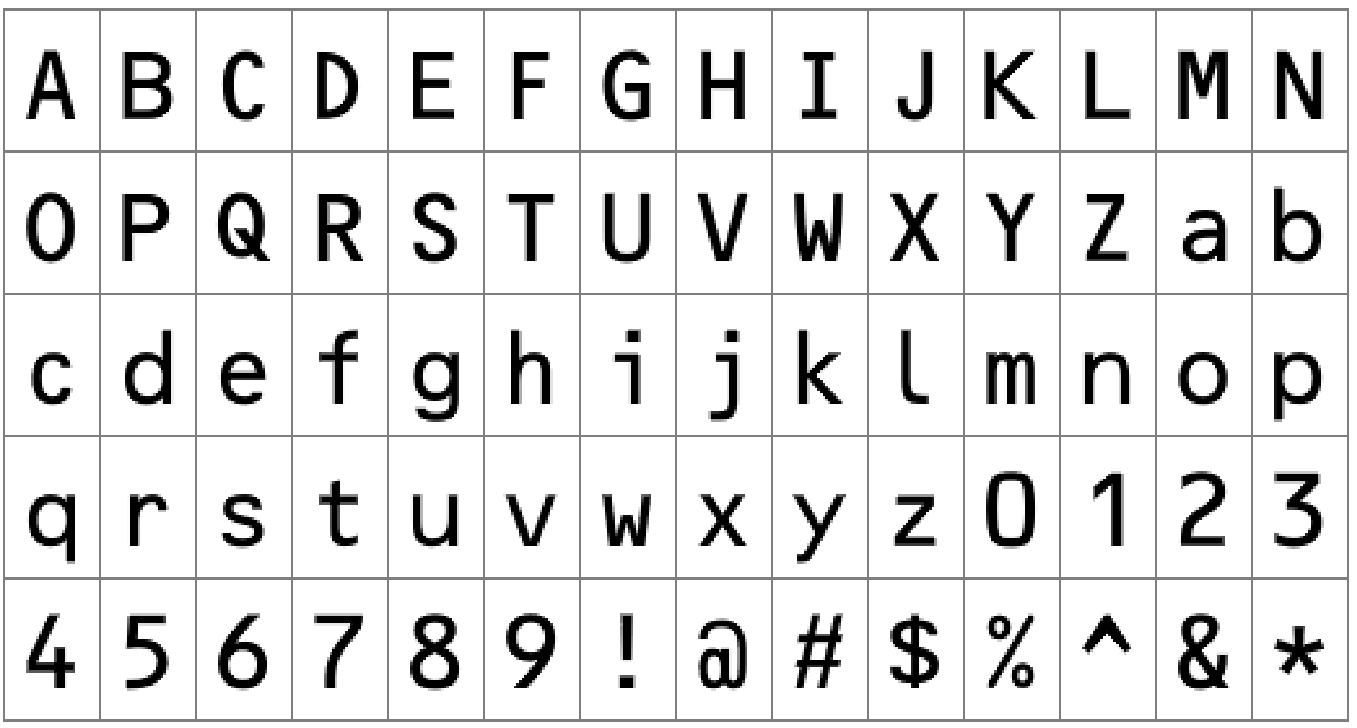}
\caption{OCR-B font}
\label{figOCRBFont}
\end{figure}

\begin{figure}
\centering
\epsfig{file=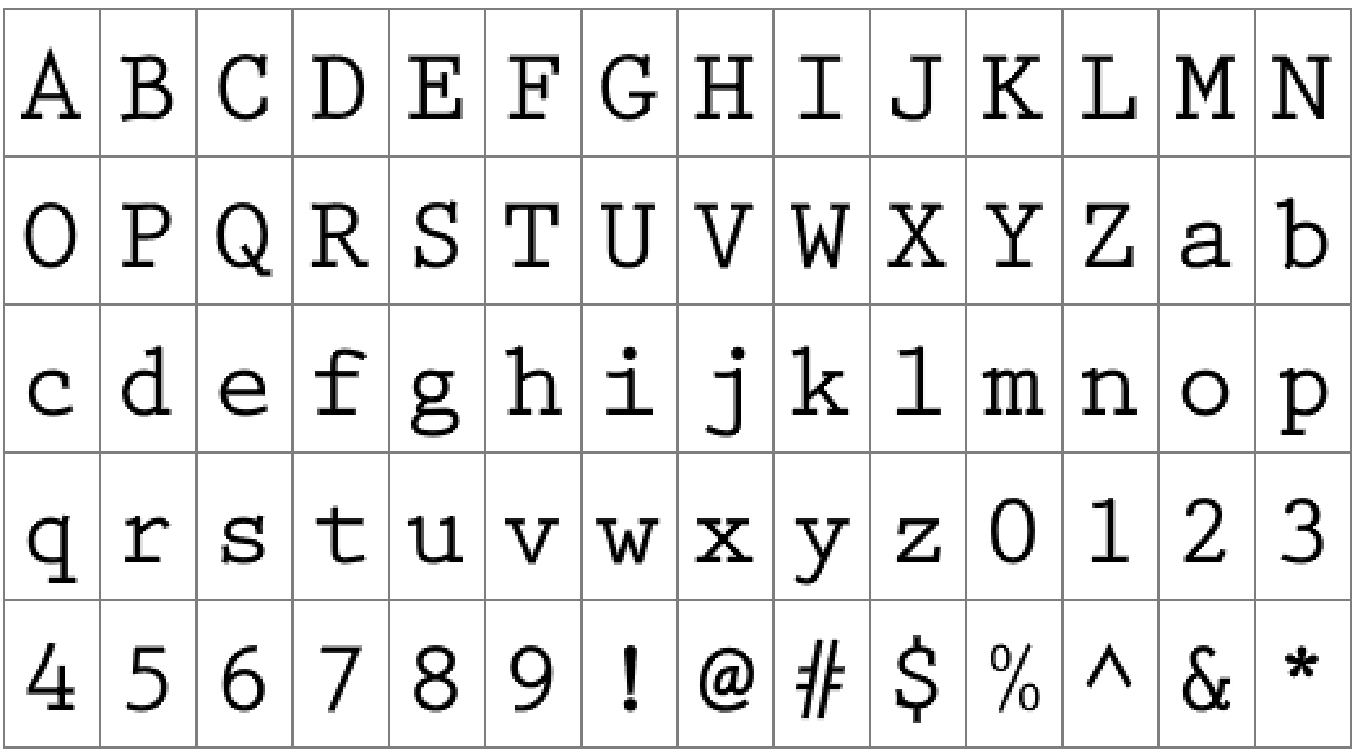}
\caption{Elite font}
\label{figEliteFont}
\end{figure}

\begin{figure}
\centering
\epsfig{file=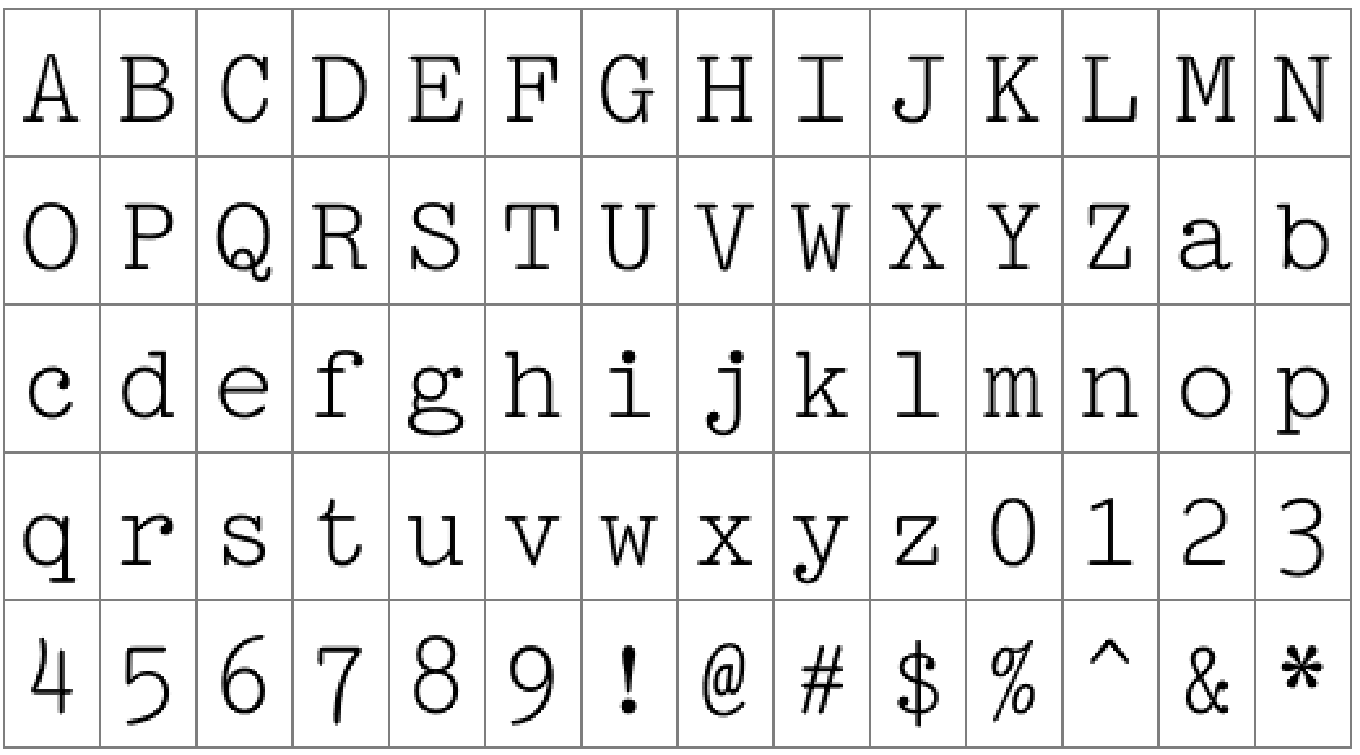}
\caption{Pica font}
\label{figPicaFont}
\end{figure}

Progress in optical character recognition has been quite spectacular since its commercial introduction in the mid 1950's. After specially-designed ``standard" typefaces, e.g. OCR-A and OCR-B (Figures \ref{figOCRAFont} and \ref{figOCRBFont} respectively), there came support for elite and pica fixed-pitch typescripts (shown in Figures \ref{figEliteFont} and \ref{figPicaFont}), and finally for omni-font typeset text (like in the document you are reading). The early experimental OCR systems were often rule-based. By the 1980's they had been completely replaced by systems based on statistical \emph{pattern recognition}. For clearly segmented printed materials using European scripts, such pattern recognition techniques offered virtually error-free OCR. Since the 1990's, the acceptance rates of form readers (usually running at a high reject/error ratio) on hand-printed digits and constrained alphanumeric fields has risen significantly. Many researchers are currently focusing on off-line and on-line cursive writing or are working on multi-lingual recognition in a variety of scripts.

When the number of symbols is large, as in the Chinese or Korean writing systems, or when the symbols are not separated from one another, as in Arabic or Devanagari print, current OCR systems are still far from the error rates of human readers, and the OCR accuracy drops even further when the source image quality is compromised e.g. by fax transmission.

In the recognition of hand-printed text, algorithms with successive segmentation, classification, and identification (language modeling) stages currently prevail. Statistical techniques based on \emph{Hidden Markov Models} (HMMs, often employed in speech recognition), have been successfully applied to recognizing cursive handwriting. HMM-based techniques make the segmentation, classification, and identification decisions in parallel, but their performance still leaves much to be desired, primarily because the inherent variability of handwriting is greater than that of speech, to the extent that dealing with illegible handwriting is more common than dealing with unintelligible speech.

Decreases in OCR systems cost have been perhaps even more striking than the improvements of the scope and accuracy in classification methods. The early OCR devices all required expensive scanners and special-purpose electronic or optical hardware, e.g. the IBM 1975 Optical Page Reader for reading typed earnings reports at the Social Security Administration cost over three million dollars. Around 1980, the introduction of affordable microprocessors for personal computers and charge-coupled array scanners resulted in a great cost decrease that paralleled that of general-purpose computers. Today, OCR software is a common add-on to any desktop scanner that costs about the same as (if not less than) a printer or a facsimile machine.

\subsection{Promising directions}
With all its challenges and difficulties, optical character recognition nevertheless is a success story in a large an exciting field of machine vision, which has known many other successful man-made object (or pattern) recognition systems. To continue this exciting success story, the OCR community has to emphasize its technological strengths, look for new ideas by investing in research and following the most promising directions that include
\begin{itemize}
\item \emph{adaptive OCR} aiming at robust handling of a wider range of printed document imagery
\item \emph{document image enhancement} as part of OCR pre-processing
\item \emph{intelligent use of context} providing a bigger picture to the OCR engine and making the recognition task more focused and robust
\item \emph{handwriting recognition} in all forms, static and dynamic, general-purpose and task-specific, etc.
\item \emph{multi-lingual OCR}, including multiple embedded scripts
\item \emph{multi-media OCR} aiming to recognize any text captured by any visual sensor in any environment
\end{itemize}
Some of the above OCR topics may sound a little ambitious, but with the current growth rates of information technology in general and OCR in particular, we believe that they are not out of reach.
\section{Acknowledgements}
This research has been carried out partly by a request from the U.S. Army Research Laboratory (ARL). This report has been created within American Management Systems, Inc. (AMS) research and development team at Lanham, MD. The authors would like to thank Ilya Zavorin, Kristen Summers, and Mark Turner of AMS for  their help in determining the main directions, formulating the scope, pointing to the important references and providing their helpful comments.
\bibliographystyle{plain}
\bibliography{report}
\end{document}